# The Unfairness of Fair Machine Learning: Levelling down and strict egalitarianism by default

Brent Mittelstadt,[1] Sandra Wachter,[2] Chris Russell[3]

## CONTENTS

[1] Corresponding author: brent.mittelstadt@oii.ox.ac.uk. Oxford Internet Institute, University of Oxford, 1 St. Giles, Oxford, OX1 3JS, United Kingdom.

[2] Oxford Internet Institute, University of Oxford, 1 St. Giles, Oxford, OX1 3JS, United Kingdom.

[3] Oxford Internet Institute, University of Oxford, 1 St. Giles, Oxford, OX1 3JS, United Kingdom. Chris Russell is also an employee of Amazon Web Services. He did not contribute to this research in his capacity as an Amazon employee. This work has been supported through research funding provided by the Wellcome Trust (grant nr 223765/Z/21/Z), Sloan Foundation (grant nr G-2021-16779), the Department of Health and Social Care (via the AI Lab at NHSx), and Luminate Group to support the Trustworthiness Auditing for AI project and Governance of Emerging Technologies research programme at the Oxford Internet Institute, University of Oxford.

ABSTRACT

In recent years fairness in machine learning (ML), artificial intelligence (AI), and algorithmic decision-making systems has emerged as a highly active area of research and development. To date, the majority of measures and methods to mitigate bias and improve fairness in algorithmic systems have been built in isolation from policy and civil societal contexts and lack serious engagement with philosophical, political, legal, and economic theories of equality and distributive justice. Most define fairness in simple terms, where fairness means reducing gaps in performance or outcomes between demographic groups while preserving as much of the accuracy of the original system as possible. This oversimplification of equality through fairness measures is troubling. Many current fairness measures suffer from both fairness and performance degradation, or "levelling down," where fairness is achieved by making every group worse off, or by bringing better performing groups down to the level of the worst off. Levelling down is a symptom of the decision to measure fairness solely in terms of equality, or disparity between groups in performance and outcomes, while ignoring other relevant features of questions of distributive justice (e.g., welfare, priority) which are more difficult to quantify and measure. When fairness can only be measured in terms of distribution of performance or outcomes, corrective actions can likewise only target how these goods are distributed between groups: we refer to this trend as 'strict egalitarianism by default'.

Strict egalitarianism by default runs counter to both the stated objectives of fairness measures as well as the presumptive aim of the field: to improve outcomes for historically disadvantaged or marginalised groups. When fairness can only be achieved by making everyone worse off in material or relational terms through injuries of stigma, loss of solidarity, unequal concern, and missed opportunities for substantive equality, something would appear to have gone wrong in translating the vague concept of 'fairness' into practice. Levelling down should be rejected in fairML because it (1) unnecessarily and arbitrarily harms advantaged groups in cases where performance is intrinsically valuable, such as medical applications of AI; (2) demonstrates a lack of equal concern for affected groups, undermines social solidarity, contributes to stigmatisation; (3) fails to live up to the substantive aims of equality law and fairML, and squanders the opportunity afforded by interest in algorithmic fairness to substantively address longstanding social inequalities; and (4) fails to meet the aims of many viable theories of distributive justice including pluralist egalitarian approaches, prioritarianism, sufficientarianism, and others.

This paper critically scrutinises these initial observations to determine how fairML can move beyond mere levelling down and strict egalitarianism by default. We examine the causes and prevalence of levelling down across fairML, and explore possible justifications and criticisms based on philosophical and legal theories of equality and distributive justice, as well as equality law jurisprudence. We find that fairML does not currently engage in the type of measurement, reporting, or analysis necessary to justify levelling down in practice. The types of decisions for which ML and AI are currently used, as well as inherent limitations on data collection and measurement, suggest levelling down can rarely be justified in practice. We propose a first step towards substantive equality in fairML: "levelling up" systems by design through enforcement of minimum acceptable harm thresholds, or "minimum rate constraints," as fairness constraints. We likewise propose an alternative harms-based framework to counter the oversimplified egalitarian framing currently dominant in the field and push future discussion more towards substantive equality opportunities and away from strict egalitarianism by default.

## 1 INTRODUCTION[4]

In recent years fairness in machine learning (ML), artificial intelligence (AI), and algorithmic decision-making systems has emerged as a highly active area of research and development. Predicted and actual uses of these technologies to distribute outcomes and resources in high-stakes domains such as medicine, law, and finance have rightly driven interest in the fairness and bias of those decisions. Deployment of these technologies has been contested by policymakers [5] and researchers [6] based on a perceived lack of trustworthiness and fairness.

The field of fair machine learning (fairML) has been predominantly driven by researchers and practitioners working in ML and AI,

---

[4] This paper would not exist without Jade Thompson, Professor Andrew Przybylski, and our amazing colleagues in the Governance of Emerging Technologies research programme at the Oxford Internet Institute. This work has been supported through research funding provided by the Wellcome Trust (grant nr 223765/Z/21/Z), Sloan Foundation (grant nr G-2021-16779), the Department of Health and Social Care (via the AI Lab at NHSx), and Luminate Group.

[5] M. WALPORT & M. SEDWILL, *Artificial intelligence: opportunities and implications for the future of decision making*, (2016), https://assets.publishing.service.gov.uk/government/uploads/system/uploads/attachment_data/file/566075/gs-16-19-artificial-intelligence-ai-report.pdf; HOUSE OF COMMONS SCIENCE AND TECHNOLOGY COMMITTEE, *The big data dilemma*, 56 (2016), https://publications.parliament.uk/pa/cm201516/cmselect/cmsctech/468/468.pdf.

[6] Nina Grgic-Hlaca et al., *Human Perceptions of Fairness in Algorithmic Decision Making: A Case Study of Criminal Risk Prediction*, in PROCEEDINGS OF THE 2018 WORLD WIDE WEB CONFERENCE 903 (2018), https://doi.org/10.1145/3178876.3186138 (last visited Jun 10, 2022); Maximilian Kasy & Rediet Abebe, *Fairness, Equality, and Power in Algorithmic Decision-Making*, in PROCEEDINGS OF THE 2021 ACM CONFERENCE ON FAIRNESS, ACCOUNTABILITY, AND TRANSPARENCY 576 (2021), https://doi.org/10.1145/3442188.3445919 (last visited Jun 10, 2022).

computer science, software engineering, and mathematics. These groups have developed numerous measures and methods to mitigate bias and improve fairness in algorithmic systems. However, the majority of these tools have been built in isolation from policy and civil societal contexts and lack serious engagement with philosophical, political, legal, and economic theories of equality and distributive justice.[7] Reflecting this, most define fairness in simple terms, where fairness means reducing gaps in performance or outcomes between demographic groups. Successfully achieving algorithmic fairness has come to mean satisfying one of these simple mathematical definitions, while preserving as much of the accuracy of the original system as possible.

This oversimplification of equality through fairness measures could possibly be attributed to the relative youth of fairML. However, the practical impact of the approach adopted by the field to date is morally troubling. Many current fairness measures have been shown to suffer from both fairness and performance degradation, or "levelling down," where fairness is achieved by making every group worse off, or by bringing better performing groups down to the level of worse performing groups.[8] Levelling down is effectively fairness achieved by breaking the system, for example by making a classifier less accurate so it performs equally badly across all relevant groups.

Levelling down is a symptom of the decision to measure fairness solely in terms of equality, or disparity between groups in performance and outcomes, while ignoring other relevant features of distributive justice such as absolute welfare or priority which are more difficult to quantify and directly measure in research and development environments. When fairness can only be measured in terms of distribution of performance or outcomes, corrective actions can likewise only target how these goods are distributed between groups. The field effectively only has egalitarian tools at its disposal which value equality of treatment and outcomes while ignoring other goods of distributive justice. Likewise, the prevalence of levelling down in fairML suggests that the field is, intentionally or otherwise, adopting a strict egalitarian approach to questions of distributive justice in which the only (measurable) value is equality. We name these trends in fairML 'strict egalitarianism by default'.

Strict egalitarianism by default, at least in its most gratuitous forms, runs counter to both the stated objectives of fairness measures as well as the presumptive aim of the field: to improve outcomes for historically disadvantaged or marginalised groups.[9] It conceives of equality in simplistic comparative terms, ignoring absolutes of welfare and justice which are necessary to achieve substantive equality rather than mere formalistic equality, or equal treatment..[10]

When fairness can only be achieved by making everyone worse off in material or relational terms through injuries of stigma, loss of solidarity, unequal concern, and missed opportunities for substantive equality, something would appear to have gone wrong in translating the vague concept of 'fairness' into practice. Equality should aim to make people better off, not simply to reduce them to a common level of harm.[11] Simple mathematical definitions can be satisfied without regard for how parity is achieved in practice and the significant material and relational harms, and opportunity costs, for the people affected. The huge interest that exists algorithmic fairness provides an opportunity to substantively address longstanding inequalities in society. Enforcing fairness solely through levelling down squanders this chance to achieve substantive rather than mere formalistic equality.

---

[9] FairML does not have universally agreed guiding principles, but prior work can provide some indication of its values and aims. In a 2019 paper critical of the state of the field, Keyes et al. defined the 'Fair' value of the Fairness, Accountability, and Transparency in Machine Learning (FAT-ML or FAccT-ML) research network as ensuring that algorithmic systems are "lacking biases which create unfair and discriminatory outcomes." See: Os Keyes, Jevan Hutson & Meredith Durbin, *A mulching proposal: Analysing and improving an algorithmic system for turning the elderly into high-nutrient slurry*, in EXTENDED ABSTRACTS OF THE 2019 CHI CONFERENCE ON HUMAN FACTORS IN COMPUTING SYSTEMS 1 (2019). More recently, Cooper et al. suggest that the field is motivated by the fact that "automated decision systems that do not account for systemic discrimination in training data end up magnifying that discrimination; to avoid this, such systems need to be proactive about being fair." See: A. Feder Cooper, Ellen Abrams & Na Na, *Emergent Unfairness in Algorithmic Fairness-Accuracy Trade-Off Research*, in PROCEEDINGS OF THE 2021 AAAI/ACM CONFERENCE ON AI, ETHICS, AND SOCIETY 46, 51 (2021), https://dl.acm.org/doi/10.1145/3461702.3462519 (last visited Jul 13, 2022). Early case studies in the field are similarly instructive, such as the famous COMPAS case in which a risk recidivism algorithm was alleged by journalists at ProPublica to be biased against Black defendants, routinely assigning them higher risk scores than comparable white defendants. See: Julia Angwin et al., *Machine bias*, 23 PROPUBLICA, MAY 2016 (2016). These examples suggest work on algorithmic fairness is motivated at least in part by a desire to improve the situation of disadvantaged people that unjustifiably receive worse treatment or outcomes than their peers. How fairness, discrimination, and bias are conceptualised and measures, and likewise what is justified in differential treatment, opportunities, and results of course differs drastically across the field and use cases, but the underlying motivation to help people who are unjustifiably harmed by algorithmic systems seems clear and uncontroversial.

[10] Sandra Wachter, Brent Mittelstadt & Chris Russell, *Bias preservation in machine learning: the legality of fairness metrics under EU non-discrimination law*, 123 W. VA. L. REV. 735 (2021).

[11] LARRY S. TEMKIN, INEQUALITY (1993); Nils Holtug, *Egalitarianism and the Levelling down Objection*, 58 ANALYSIS 166 (1998); Brett Doran, *Reconsidering the Levelling-down Objection against Egalitarianism*, 13 UTILITAS 65 (2001); Derek Parfit, *Equality or Priority?*, in THE IDEAL OF EQUALITY 81 (Matthew Clayton & Andrew Williams eds., 2002).

This paper critically scrutinises these initial observations to determine whether, and to what extent, levelling down and strict egalitarianism by default are problematic for fairML. Sections 2 and 3 introduce the concept of levelling down and examines its prevalence across fairML research and development. Section 4 draws on philosophical and legal theories of equality and distributive justice, as well as equality law jurisprudence, to explore possible justifications and criticisms of levelling down as a tool of distributive justice. Section 5 then considers the relevance and feasibility of these possible justifications in the context of fairML, concluding that the field does not currently engage in the type of thinking necessary to justify equality achieved through levelling down. The types of decisions for which ML and AI are currently used, as well as inherent limitations on data collection and measurement, both suggest levelling down can rarely be justified in practice. Section 6 describes an alternative approach to fairness by "levelling up" systems by design through enforcing minimum acceptable harm thresholds, or "minimum rate constraints," as fairness constraints. We propose an alternative harms-based framework to counter the oversimplified egalitarian framing currently dominant in the field and push future discussion more towards substantive equality opportunities. Section 7 concludes with recommendations for a normative and practical reconfiguration of the field to resist levelling down and strict egalitarianism by default.

## 2   LEVELLING DOWN

Methods which enforce group-based parity measures of fairness (or 'group fairness' measures) are morally and legally problematic due to (1) their implicit values, which disproportionately favour equality over reduction of harm, and (2) the way they achieve equality in practice through 'levelling down', by which certain groups are needlessly made worse off for the sake of mathematical convenience. Far from being a mathematical or theoretical exercise, the enforcement of fairness in these terms arbitrarily harms people subject to decisions informed by 'fair' ML or AI systems.[12]

Broadly we can decompose the machinery of fair machine learning (ML) into two components: measures and methods. Measures are simple computations, such as the difference in true positive rate between two protected groups, that describe how unfair the behaviour of a system is or appears to be. Methods are novel approaches to ML that improve these measures, equalising rates of harm at the expense of criteria such as overall accuracy, which are typically optimised by an ML system trained without consideration of fairness.

---

[12] For example, in the case of hiring, lower hiring rates, or in the case of cancer detection, an increased failure to identify people who have cancer as having cancer.

When we build ML systems to make decisions about people's lives, our design decisions encode implicit value judgments about what properties should be prioritized by the system's behaviour. For example, standard ML systems are typically trained to maximise some notion of accuracy by minimizing a proxy such as log loss.[13] Methods used to enforce fairness in ML systems, or "algorithmic fairness methods," likewise impose certain value judgements about which properties a system should optimize, for example valuing equality of error rates over accuracy, and alter system behaviour accordingly.

The idea that accuracy is not always the most relevant property for evaluating performance of a model is commonly accepted across ML research. For example, when dealing with rare events, such as trying to identify forms of cancer that occur in less than 1% of the population, a constant classifier that always predicts that cancer is not present will have over 99% accuracy. It may likewise have higher accuracy than other models or classification methods that would, nonetheless, be more useful in practice.

In such cases involving severely unbalanced datasets, properties such as precision (i.e., the proportion of people identified as being at risk of cancer that actually have cancer) or recall (i.e., the proportion of the people who will eventually develop cancer that are correctly identified as being at risk of having cancer) are often more useful. Positive decision rates may also be a more relevant property. In the case of rare cancer screening the proportion of people diagnosed as being at risk may also be a more relevant property to optimize. For example, a healthcare authority may only have resources to screen at most $k\%$ of the population, in which case they may prefer a model that maximises recall while keeping the percentage of the population called for screening under this $k\%$.

## 2.1   Levelling down via group fairness

Standard group-based parity measures of fairness, or 'group fairness', tend to achieve fairness by selecting one or more properties that are more important than accuracy for a particular case, and then enforcing equality for this property across relevant demographic groups while preserving accuracy as far as possible. Example measures include equality of accuracy, equal opportunity (corresponding to equality of

---

recall across demographic groups), equality of precision, and demographic parity (corresponding to equality of positive rate).[14]

Once a property has been selected, equality can be enforced in two ways: (1) adjust performance along the chosen property for the disadvantaged group, for example by improving recall at the cost of accuracy, and (2) degrade performance for advantaged groups along the same property. These approaches tend to be combined in practice to satisfy group fairness measures as fully as possible.

Concerning the former, enforcing equality for one or more properties while also maximising accuracy often requires altering the behaviour of a classifier for multiple groups.[15] For example, groups which have a below average positive decision rate or recall, henceforth referred to as "disadvantaged groups," can have these properties increased by enforcing a group fairness method on the classifier. Gains in positive decision rates or recall typically come at the cost of accuracy. Nonetheless, if groups are thought to be harmed by low decision rates or low recall, this trade-off can be considered beneficial or justified.

Concerning the latter, group fairness measures also tend to alter performance for groups with above average performance, henceforth referred to as "advantaged groups." Performance for relevant properties, such as decision rate or recall, tend to be degraded. This degradation likewise comes at the cost of accuracy.[16] Assuming that performance measures like higher decision rates or recall are inherently valuable properties, and higher accuracy is likewise valuable, the resulting fair classifier would be Pareto inefficient.[17]

Enforcing equality by degrading performance for advantaged groups causes the phenomenon we refer to as "levelling down." Performance is arbitrarily degraded for advantaged group(s) solely to reduce disparity in a given property (e.g., recall, decision rates) between groups while maintaining as much accuracy as possible. Levelling down occurs where equality is enforced not only by increasing the relevant property for disadvantaged groups, but by arbitrarily making one or more better performing groups worse off by reducing performance for them along the same property.

Here, we are concerned with cases where the degradation of performance for advantaged groups is not causally linked to improvements in performance for disadvantaged groups. In such cases the loss of performance for advantaged groups is not strictly necessary to improve recall, decision rates, or some other valuable property for disadvantaged groups; rather, it is inflicted solely to reduce performance disparity in the chosen property, thereby satisfying a mathematical definition of fairness which says a model is fair when this number is equal between groups.

---

[14] Sahil Verma & Julia Rubin, *Fairness definitions explained*, *in* 2018 IEEE/ACM INTERNATIONAL WORKSHOP ON SOFTWARE FAIRNESS (FAIRWARE) 1 (2018).

[15] Sam Corbett-Davies et al., *Algorithmic decision making and the cost of fairness*, *in* PROCEEDINGS OF THE 23RD ACM SIGKDD INTERNATIONAL CONFERENCE ON KNOWLEDGE DISCOVERY AND DATA MINING 797 (2017).

[16] *Id.*

[17] ZIETLOW ET AL., *supra* note 8.

Our concern arises because this behaviour introduces additional and avoidable harm to better performing groups solely to achieve parity between groups. Levelling down does not directly benefit worse performing groups. A more instructive framing to capture this behaviour is to think of the choice of group fairness measures as a choice about the type of harm that should be equalised between groups. A table examining the type of harm equalised by different group fairness measures is available in Appendix 1.[18]

Equal opportunity, for example, requires recall rates to be equalised between groups while maintaining as much accuracy as possible. The harm caused is a loss of recall for better performing groups, or a greater failure to correctly identify positive cases. A mathematically optimal solution that equalises recall at a minimum loss to accuracy would involve both steps mentioned above: increase recall for worse performing groups, but also reduce it for better performing groups.[19] If enforced on a cancer screening system, for example, equalising recall means that more cases of cancer will be missed for better performing groups than would have otherwise been the case. What is not clear, and what we will examine throughout this paper, is whether the avoidable harms for both advantaged and disadvantaged groups which are caused by levelling down can be ethically, legally, and socially justifiable.

## 2.2   Example: levelling down in cancer screening

As an illustrative example of the harm of levelling down in practice, imagine that we want to enforce fairness in AI system used for predicting future risk of lung cancer. Our imaginary system, which is inspired by a real-world patient triage system,[20] suffers from a performance gap between Black and white patients. Specifically, the system has lower recall for Black patients, meaning it routinely underestimates their risk of cancer and incorrectly classifies patients as "low risk" who will eventually develop lung cancer.

---

[18] In no small part, the wide range of fairness measures now available in fairML can be attributed to researchers identifying relevant properties from the classification literature and then finding ways to enforce equality with respect to these properties. See: Verma and Rubin, *supra* note 14.

[19] Corbett-Davies et al., *supra* note 15.

[20] Ziad Obermeyer & Sendhil Mullainathan, *Dissecting racial bias in an algorithm that guides health decisions for 70 million people, in* PROCEEDINGS OF THE CONFERENCE ON FAIRNESS, ACCOUNTABILITY, AND TRANSPARENCY 89 (2019).

The worse baseline performance can have many causes. It may have resulted from our system being trained on data predominantly taken from white patients, or because health records from Black patients are less accessible or lower quality. Likewise, it may reflect underlying worse performance in existing clinical technologies, or social inequalities in healthcare access and expenditures. Racially biased medical devices, for example, caused delayed treatment for darker skinned patients during the COVID-19 pandemic because pulse oximeters overestimated blood oxygen levels in minorities.[21] Similarly, lung and skin cancer detection technologies have been shown to be less accurate for darker skinned people meaning they more frequently fail to flag cancers in patients, delaying access to life saving care.[22] Patient triage systems regularly underestimate the need for care in minority ethnic patients, for example due to using health care costs as a proxy for illness while failing to account for unequal access to care, and thus unequal costs, across the population.[23][24]

Whatever the cause of the performance gap, our motivation for enforcing fairness is to substantively improve performance for a worse performing (or 'disadvantaged') group, in this case Black patients. In the context of cancer screening, false negatives are much more harmful than false positives; the latter mean that the patient will have unnecessary health checks or scans, whereas the former means their more future cases of cancer will go undiagnosed and untreated.

One way to improve the situation of Black patients is therefore to improve the system's recall. As a first step we may decide to err on the side of caution and tell the system to change its predictions for the cases it is least confident about involving Black patients. Specifically, we would flip some low confidence "low risk" cases to "high risk" in order to catch more cases of cancer in the future. This change comes at the cost of accuracy; the number of people incorrectly classified as the being at risk of cancer goes up, and the system's overall accuracy goes down. However, this trade-off between accuracy and recall is acceptable

---

[21] Ashraf Fawzy et al., *Racial and Ethnic Discrepancy in Pulse Oximetry and Delayed Identification of Treatment Eligibility Among Patients With COVID-19*, 182 JAMA INTERNAL MEDICINE 730 (2022).

[22] David Wen et al., *Characteristics of publicly available skin cancer image datasets: a systematic review*, 4 THE LANCET DIGITAL HEALTH e64 (2022).

[23] Obermeyer and Mullainathan, *supra* note 20.

[24] Similar bias can also be observed along gender lines, with female patients being disproportionately misdiagnosed and mistreated for heart disease. See: Nancy N. Maserejian et al., *Disparities in Physicians' Interpretations of Heart Disease Symptoms by Patient Gender: Results of a Video Vignette Factorial Experiment*, 18 J WOMENS HEALTH (LARCHMT) 1661 (2009); Linda Worrall-Carter et al., *Systematic review of cardiovascular disease in women: Assessing the risk*, 13 NURSING & HEALTH SCIENCES 529 (2011).

because failing to predict a future case of cancer can severely harm patients.

By flipping cases in this way to increase recall at the cost of accuracy we can eventually reach a state where any further changes would come at an unacceptably high loss of accuracy. Where this threshold lies in practice is ultimately a subjective decision; there is no objective tipping point between recall and accuracy. We have not necessarily brought performance (or recall) for Black patients up to the same level as white patients, but we have done as much as possible within the constraints of the current system and available data and other resources to improve the situation of Black patients and reduce the performance gap.

This is where we encounter the key dilemma responsible for levelling down in fairML. We can take an optional second step to further reduce the performance gap between Black and white patients. We cannot improve performance for Black patients any further without an unacceptable loss of accuracy. However, we can still reduce performance for white patients, lowering both their recall and accuracy in the process, so that our system performs equally well, or as close as possible, for both groups.

This is what many group fairness methods do in practice. The motivation is mathematical convenience: the aim is to make two numbers (i.e., recall) as close to equal as possible between two groups (i.e., white and Black patients), solely to satisfy a mathematical definition that says a system is fair when these two numbers are equal. In our example, we would alter the labels of white patients as well, switching some of the predictions from high risk to low risk.

Clearly, this type of label flipping for the sake of equality can be extremely harmful for patients who would not be offered follow-up care and monitoring. Overall accuracy decreases and the frequency of the most harmful type of error increases, all for the sake of reducing the gap in performance. In our example, this levelling down does not improve the situation of Black patients (who already have a classifier with improved recall); rather, it serves only to equal out performance (or recall) between Black and white patients. It can likewise cause broader social harms, undermine more difficult but substantively rich solutions to inequality (e.g., increased access to healthcare, improved data quality), and cause stigmatisation and social isolation (see: Section 4).

Levelling down thus benefits neither group directly. Assuming a system has already been designed to minimise costs for all patients, it would be inappropriate to choose an intervention that "would inevitably make at least one group worse off without making the other group better off,"[25] and yet this is precisely what current applications of group fairness achieve in fairML.

## 3    HOW COMMON IS LEVELLING DOWN IN FAIRML?

The usage of equality-based fairness measures in machine learning is not self-evidently troubling; rather, it is how they are enforced in practice, and the resulting levelling down, which causes problems. When used solely for diagnostic purposes, egalitarian measures such as (conditional) demographic parity [26] or equal opportunity [27] provide a helpful warning that different groups are being treated differently, and that they are potentially being harmed in different ways by an algorithmic decision-making system. However, using them to determine which models should be deployed in real world use cases raises serious ethical and legal concerns. Levelling down can occur as a direct result of the use of egalitarian measures in model selection. In many ways this problem is another example of Goodhart's Law that "when a measure becomes a target, it ceases to be a good measure."[28]

   In this section we demonstrate how levelling down occurs for a range of fairness measures, focusing on two of the most commonly used metrics in the fairML literature: demographic parity and equal opportunity. To do so, we enforce these measures across a range of algorithms using two of the most widely used fairness toolkits: FairLearn and IBM AI Fairness 360 (IBM360).[29] We show that the existence of levelling down is not a limitation or design flaw of these toolkits or methodologies; rather, it is a natural consequence of strictly enforcing equality as part of model selection.[30] As such, levelling down

---

[25] Chloé Bakalar et al., *Fairness on the ground: Applying algorithmic fairness approaches to production systems*, ARXIV PREPRINT ARXIV:2103.06172, 5 (2021).

[26] Sandra Wachter, Brent Mittelstadt & Chris Russell, *Why fairness cannot be automated: Bridging the gap between EU non-discrimination law and AI*, 41 COMPUTER LAW & SECURITY REVIEW 105567 (2021); Faisal Kamiran & Toon Calders, *Data preprocessing techniques for classification without discrimination*, 33 KNOWLEDGE AND INFORMATION SYSTEMS 1 (2012).

[27] Moritz Hardt, Eric Price & Nati Srebro, *Equality of opportunity in supervised learning, in* ADVANCES IN NEURAL INFORMATION PROCESSING SYSTEMS 3315 (2016).

[28] Marilyn Strathern, *'Improving ratings': audit in the British University system*, 5 EUROPEAN REVIEW 305 (1997).

[29] Rachel KE Bellamy et al., *AI Fairness 360: An extensible toolkit for detecting and mitigating algorithmic bias*, 63 IBM JOURNAL OF RESEARCH AND DEVELOPMENT 4: 1 (2019); Sarah Bird et al., *Fairlearn: A toolkit for assessing and improving fairness in AI*, MICROSOFT, TECH. REP. MSR-TR-2020-32 (2020).

[30] For example, Kim discusses a range of changes to the design of an ML algorithm that could decrease the "disparate impact" (a concept from US anti-discrimination law loosely corresponding to demographic parity) of the decisions made by a system. See: Pauline Kim, *Race-Aware Algorithms: Fairness, Nondiscrimination and Affirmative Action*, CALIFORNIA LAW REVIEW (2022), https://papers.ssrn.com/abstract=4018414 (last visited Jul 13, 2022). If a data scientist systematically explored combinations of these changes, and then selected the model with disparate impact below a predetermined

is a concern for anyone that takes group fairness measures into consideration when deciding which model should be deployed, and not just those who try to enforce fairness through ML toolkits.

## 3.1   Why does levelling down occur?

To understand why levelling down occurs it is vital to recognise that most notions of group fairness are underspecified. To enforce demographic parity, for example, it is only necessary to create a classifier that gives positive decisions to 100% of people, or to 0% of people, or to any ratio in between, providing that the same proportion of each group (e.g. Black and white patients) receive positive decisions. The same ambiguity occurs when enforcing equal opportunity, only here the recall of each group must be matched rather than the ratio of positive decisions.[31]

Given this ambiguity, how are models actually selected using group fairness measures? In practice, models are selected to maximize some notion of classification performance such as accuracy, balanced accuracy, F1-score, or Matthew's Correlation Coefficients (MCC),[32] while also being sufficiently equal to satisfy a chosen fairness measure as far as possible. Reflecting this, comparisons between models on the basis of fairness and accuracy are common in the ML literature. Methods are competitively benchmarked based on their ability to obtain higher levels of accuracy for a given level of fairness.[33]

It is this combination of maximizing accuracy while enforcing equality that leads to levelling down. To understand why, it is necessary to briefly discuss how ML classifiers operate.

ML classifiers tend not only assign labels to datapoints corresponding to individuals, but also to assign a notion of confidence indicating how likely it is that a particular label is correct. Low confidence datapoints (e.g., an individual patient's predicted cancer risk) are least likely to be labelled correctly. Altering low confidence cases can substantially change an equality-based fairness measure

---

threshold, and that otherwise maximizes accuracy, it is likely that this would also exhibit levelling down.

[31] More generally this ambiguity holds for any notion of group equality. If we want the groups to be (approximately) equal with respect to some property such as selection rate, recall, or precision, this still leaves open the question as to what specific value should the property take.

[32] Yasen Jiao & Pufeng Du, *Performance measures in evaluating machine learning based bioinformatics predictors for classifications*, 4 QUANTITATIVE BIOLOGY 320 (2016).

[33] Model benchmarking reflects an underlying truth about how ML is used in practice: it is used precisely because some notion of classifier performance is held to be important in a given decision-making process. If this was not the case outcomes could instead be assigned arbitrarily, and there would be no need for ML. This is not to say that performance alone is sufficient. Fairness methods are used in order to maximize performance while satisfying other criteria.

(e.g., recall rate) while having little impact on the classifier's overall accuracy. As such, in order to enforce equality while maximizing accuracy, it is beneficial to alter low confidence cases in both groups because these have the lowest "cost" in terms of accuracy.

This approach brings the groups closer to parity along a given equality measure, for example recall rate, by increasing performance (or recall) for disadvantaged groups and reducing performing (or recall) for advantaged groups. The alternate strategy of levelling up (see: Section 6), and simply increasing performance for disadvantaged groups until they obtain parity with the most advantaged groups, requires labels to be altered for datapoints where the classifier is more confident, resulting in a greater drop in accuracy. This phenomenon was formalised in a 2017 paper by Corbett-Davies et al. that showed that if a classifier is well-calibrated[34] then a greedy strategy that systematically disadvantages already advantaged groups, and advantages already disadvantaged groups, is provably optimal.[35]

However, this process of systematically disadvantaging some groups while advantaging others is optimal in a much wider range of scenarios. Whenever a classifier uses data about some set of individuals that is intrinsically uninformative, the classifier will exhibit poor accuracy for these and similar individuals. Decisions about these individuals can therefore be altered with minimal loss of accuracy. Disadvantaging these "difficult to label" individuals in advantaged groups, and likewise advantaging difficult to label individuals in disadvantaged groups, allows for substantial reductions of inequality with relatively little loss of accuracy.

While only the post-processing method described by Corbett-Davies et al. explicitly performs this type of "levelling down" to achieve equality between groups,[36] other methods for enforcing fairness in ML implicitly behave the same way. Lohaus, for example, found that a range of methods for enforcing demographic parity exhibit the same trade-off between equality and accuracy, and make statistically indistinguishable decisions about individuals.[37]

As such, levelling down is often an optimal solution to satisfy a fairness measure while retaining as much accuracy as possible. Enforcing group fairness need not, however, always result in levelling down. There are at least two broad cases in which levelling down will not occur: (1) when model selection fails, and (2) when models are not expressive enough to treat different groups differently.

---

[34] A classifier is *well-calibrated* if, for every group, the confidence score it returns corresponds to the probability that the classifier is correct.

[35] Corbett-Davies et al., *supra* note 15.

[36] *Id.*

[37] Michael Lohaus et al., *Are Two Heads the Same as One? Identifying Disparate Treatment in Fair Neural Networks*, ARXIV PREPRINT ARXIV:2204.04440 (2022).

Concerning model selection failure, Zietlow et al. analysed a range of published approaches for bias-preserving fairness in computer vision (i.e., methods that claimed to be equalizing some form of error rate between groups),[38] and observed that none used held-out data to determine error rates.[39] This failure to use held-out data is troubling in computer vision where the usage of high-capacity models means training error goes to zero, meaning the only way to reliably estimate error rates is using held-out data. As such, while these approaches did show a decrease in both accuracy and inequality, it is likely that this was due to a general deterioration in model performance, and not because fairer models were explicitly selected.[40]

Concerning inexpressive models, in the description above about how fair classifiers that maximise accuracy should behave, it is assumed that "difficult to label" individuals can be easily identified, and that their group membership can be inferred. In some cases, this may not be true. If classifiers do not have access to data about group membership, it may not be possible to infer group membership reliably enough to differentiate treatment or "flip" labels in an informed way.

In both of these cases, the behaviour is often even more concerning than levelling down. In the standard case discussed above we can be confident that at least one disadvantaged group is better off, and that the measure we are trying to equalise (e.g., recall or selection rate) has improved. In cases of model selection failure and inexpressive models, we cannot be confident in this knowledge. While a method for enforcing group fairness must result in lower inequality to be considered a success, it may do this by decreasing or increasing performance for every group, and there is no guarantee as to how this will be accomplished in practice. For example, Zietlow et al. found that most fairness methods in computer vision improved equal opportunity by decreasing the average recall for every group across a range of tasks.[41] In many high-risk situations, such as medical testing, the use of methods that improve equality by decreasing performance (e.g., diagnosis rates) for everyone would be grossly inappropriate.

## 3.2   Levelling down in practice

To further establish the prevalence of levelling down in fairML, we demonstrate its occurrence using two popular fairness toolkits using real-world code bases. Specifically, we illustrate how levelling down occurs using standard examples taken from the "How To" guides for FairLearn and IBM AI Fairness 360.

---

[38] Wachter, Mittelstadt, and Russell, *supra* note 10.

[39] Zietlow et al., *supra* note 8.

[40] *Id.*

[41] *Id.*

For Fairlearn, we ran the code from their quick start guide.[42] The code makes use of the exponentiated gradient algorithm of Agarwal et al. [43] to enforce fairness using the decision tree implementation of scikit-learn.[44] Results are shown on the UCI Adult Dataset.[45]

We make two minor modifications to the instructions and code provided in the FairLearn quick start guide. We run the code three times, enforcing all of the group fairness metrics supported by the toolkit: demographic parity, True Negative rate, and True Positive Rate, the latter of which corresponds to equal opportunity (or equal recall).[46] We also increase the maximum tree depth from 4 to 10 to make the models sufficiently expressive to differentiate between groups. Results of this experiment are displayed in Figure 1.

As is apparent in the figures, enforcing fairness through demographic parity, True Negative Rate, and True Positive Rate is achieved by increasing performance for the disadvantaged group and reducing performance for the advantaged group (i.e., 'Male' for demographic parity and True Positive Rate, 'Female' for True Negative Rate).

Similar behaviour is observed with IBM AI Fairness 360. Compared to FairLearn, IBM's toolkit is less cohesive because it is comprised of a collection of different pieces of research code written by many different authors. Surprisingly, some of the code samples provided in the toolkit fail to improve fairness on their own training set. Figure 2 displays results for one code sample that does not suffer from this weakness: reweighting pre-processing [47] on the UCI German Credit Dataset.[48]

Again, parity was achieved by both increasing performance for the disadvantaged group (i.e., 'Female') and decreasing performance for the advantaged group (i.e., 'Male'). However, despite the obvious decreases in performance for some groups in the above examples, the default reporting standards in fairML make it impossible to identify such drops. Fairness reporting tends to reduce what should be at least two measures of group performance (i.e., selection rate or recall per group) into one measure (i.e., a measure of the inequality such as difference in the selection rate or recall between groups). This simplification makes

---

it impossible to determine who is harmed, and who if anyone is helped by the enforcement of group fairness.

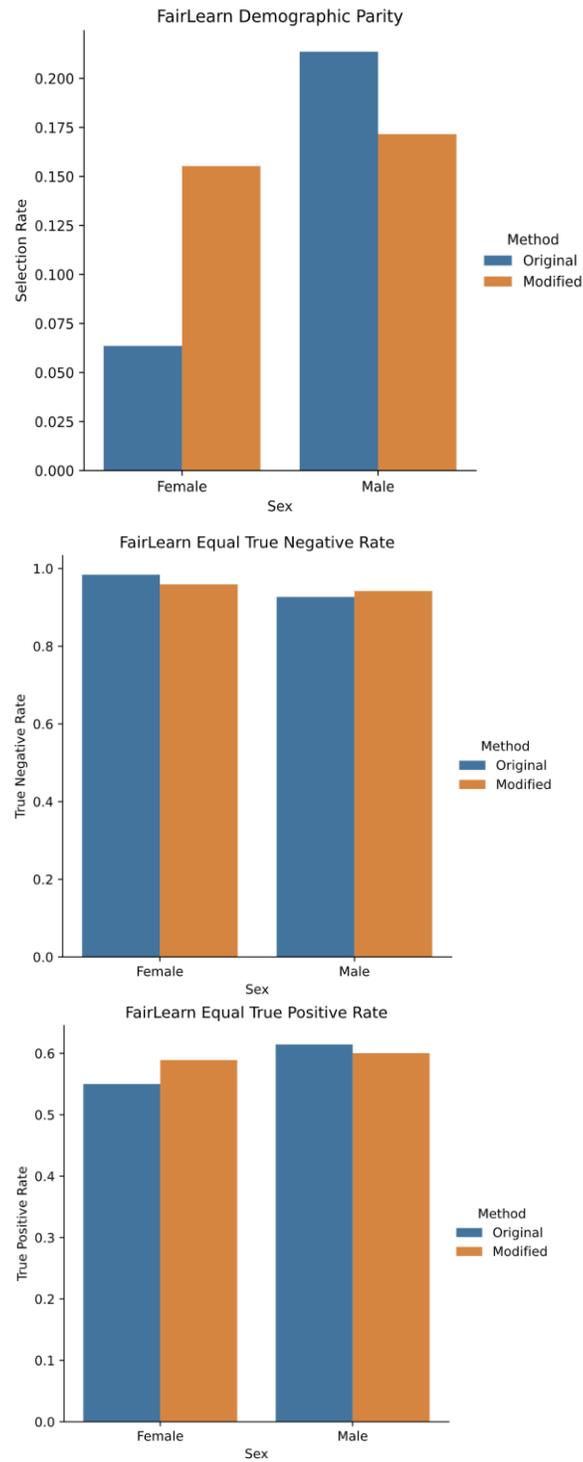

*Figure 1 - Levelling down in FairLearn on the UCI Adult Dataset*

However, even after identifying the drop in performance, it remains an open question, then, whether the phenomena observed above are genuine cases of levelling down.

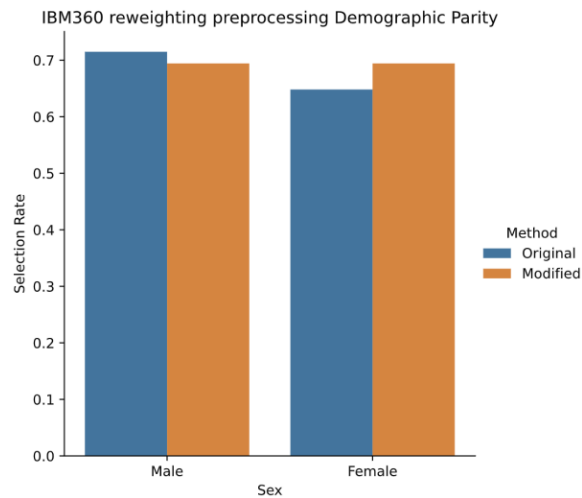

*Figure 2 - Levelling down in IBM AI Fairness 360 on the UCI Adult Dataset*

The field's focus on minimizing inequality while maximizing accuracy has left us without the tools needed to achieve fairness purely by levelling up, which would mitigate avoidable material harms, stigmatisation and loss of solidarity for both advantaged and disadvantaged groups (see: Section 4). Simply put, given how fairness is currently enforced and reported in ML, we cannot determine if harming particular groups is in fact necessary and justified, or merely the path of least resistance to achieve parity. We will definitively answer this in Section 6 where we propose new levelling up tools for algorithmic fairness and show that they can reduce harms and improve performance for disadvantaged groups without disadvantaging others.

## 3.3   Levelling down in theory

These reporting limitations make it difficult to determine the frequency and justifiability of levelling down in fairML. However, an alternative approach is to determine whether the theoretical foundations of popular fairness measures consider levelling down to be a legitimate distribution mechanism. This line of inquiry cannot, of course, establish its empirical prevalence, but can at least indicate whether levelling down is a theoretically coherent course of action to enforce group fairness measures.

Fairness measures are implicitly inspired by, or explicitly derived from, theories of distributive justice.[49] Distributive justice concerns the "relative impact of allocations on different social groups or subgroups within the population, given existing social inequalities."[50] Theories of distributive justice specify how goods and burdens should be distributed among individuals and groups in a society.[51] They tend to address allocation of resources that are rivalrous, meaning they are 'consumed' by allocation and unavailable for further distribution, and scarce, meaning there may not exist an ideal amount of the resource to satisfy everyone.[52] Justice can be conceived of in comparative or absolute terms, meaning we may be concerned simply because people are treated differently or unequally, or alternatively because their current treatment does not provide them with the basic goods they deserve (e.g., a minimum level of welfare, human rights).[53]

Group fairness measures are related to egalitarian thinking in distributive justice. Egalitarian theories assign some value to equality itself.[54] Justice is treated as a comparative concept, meaning it should be achieved through equality or reducing disparity in the distribution of a given property or resource. Group fairness measures similarly aim to ensure "some form of statistical parity (e.g. between positive outcomes, or errors) for members of different protected groups (e.g.

---

[49] Binns, *supra* note 7; Matthias Kuppler et al., *Distributive Justice and Fairness Metrics in Automated Decision-making: How Much Overlap Is There?*, (2021), http://arxiv.org/abs/2105.01441 (last visited Jul 13, 2022).

[50] Hoda Heidari et al., *On Modeling Human Perceptions of Allocation Policies with Uncertain Outcomes*, 4 (2021), http://arxiv.org/abs/2103.05827 (last visited Aug 14, 2022).

[51] Considering group fairness in ML on the one hand and distributive justice on the other necessarily leads to some terminological confusion. As explained above (see: Section 2) we refer to advantaged and disadvantaged groups in ML according to their comparative level of performance. These terms are also used in the distributive justice literature alongside terms such as "better off" and "worse off." The terms are related but distinct. (Dis)advantage in distributive justice can be understood in comparative or absolute terms and is measured by access to some good or benefit, and according to one's theoretical commitments may focus solely on distributions in a case at hand, or instead account for historical distributions and social factors which affect the relative value of goods for groups. In practice, these two uses of the terms overlap in practice; historically disadvantaged groups are often the groups which likewise suffer from worse performance in classification problems Obermeyer and Mullainathan, *supra* note 20.. We have discussed this observation, that historical inequality is a significant consideration in questions of distributive justice and should be accounted or in fairML, at length elsewhere. See: Wachter, Mittelstadt, and Russell, *supra* note 10. To avoid terminological confusion, we add the prefix "(historical)" whenever discussing historically (dis)advantaged groups in the context of ML problems.

[52] Kuppler et al., *supra* note 49; Derek Parfit, *Equality and Priority*, 10 RATIO 202 (1997).

[53] KASPER LIPPERT-RASMUSSEN, BORN FREE AND EQUAL? A PHILOSOPHICAL INQUIRY INTO THE NATURE OF DISCRIMINATION (2014).

[54] TEMKIN, *supra* note 11.

gender or race),"[55] where groups are defined by "different values for a set of protected attributes."[56] While all group fairness measures are based on some form of statistical parity, they differ in terms of their target properties, which include equality of opportunities, outcomes, treatment, mistreatment, and minimal thresholds of discrimination, among others.[57][58] Despite these differences, all group fairness measures share an egalitarian aim to achieve parity between groups along one or more chosen properties or performance measures.

In distributive justice, equality can often only be achieved by making some groups worse off [59] (see: Section 4). Strict approaches to egalitarianism, which value equality intrinsically and ignore other considerations such as welfare, view levelling down as justifiable. In other words, according to strict egalitarianism, it is acceptable to make a group worse off without directly benefiting others in order to eliminate disparity.[60] This approach views justice strictly in comparative terms and ignores absolute entitlements. Achieving equality by making everyone worse off in absolute terms is acceptable for strict egalitarians [61] despite the fact that no individual experiences a direct benefit from equality.[62] It follows that levelling down, while intuitively problematic (see: Section 4.1), is not theoretically incoherent from the view of strict egalitarianism. Methods to enforce group fairness measures based on strict egalitarianism (purposefully or otherwise) can therefore be expected to level down in at least some cases.

But how widespread are group fairness measures that align with strict egalitarianism? The majority of uses of the term 'fairness' in fairML are actually placeholders "for a variety of normative egalitarian considerations."[63] This is, however, arguably a result of how fairness is conceptualised and measured rather than explicit theoretical choices.[64]

Works in fairML that propose (group) fairness measures tend not to link them explicitly to theories of distributive justice, or to do so in a superficial manner without accounting for contextual factors or offering normative justification.[65] Others that engage seriously with distributive justice have explicitly endorsed strict egalitarianism, having deemed an "adequate justification for an unequal distribution of prediction errors" impossible for anyone to make in fairML.[66] Nonetheless, measures which conflate fairness with a strict notion of equality and equal treatment currently dominate the fairML literature.[67] Unsurprisingly, a tendency to achieve equality through levelling down has been observed for group fairness measures [68] regardless of their specific egalitarian theoretical grounding.[69]

It would seem, then, that enforcing group fairness endorses strict egalitarianism, albeit inadvertently due to how parity is measured rather than any purposeful theoretical choice. From the perspective of distributive justice this tendency is highly concerning.

## 4    IS LEVELLING DOWN JUSTIFIABLE?

While it is clear that achieving parity in performance between groups will often involve levelling down, the failure to engage in serious theoretical discussion or offer normative justifications for its necessity and justifiability in specific contexts and cases means its ethical, legal, and social acceptability remain unproven. Purposefully or otherwise, the default adoption of (strict) egalitarianism has led the field to a point where enforcement of group fairness creates avoidable harms for everyone involved.

Outside of fairML levelling down is not a new phenomenon. Philosophy has long debated the merit of theories of distributive justice. In moral philosophy the 'Levelling Down Objection' has been advanced against a strict egalitarian approach to distributive justice.[70] Strict egalitarianism measures justice solely in terms of equality. Equal states are preferred to unequal states. This is a strictly comparative approach to distributive justice: all that matters is whether the good or resource in question is equally distributed between individuals or groups independent of other considerations, such as the absolute welfare of the groups in question. Reflecting this, inequality can be reduced in two ways: either by improving the situation of groups with lower performance, or by reducing the level of all groups else to be closer to the level of the worst performing group. The latter of these two options has been coined 'levelling down'.[71] According to the levelling down objection, a strict egalitarian whose conception of justice is based solely on a comparative notion of equality would favour a state in which all people are made equally worse off in terms of welfare to one in which different people have different levels of welfare.[72][73] Strict egalitarians ignore welfare; parity is the only morally relevant consideration in distributive justice.

---

[65] Cooper, Abrams, and Na, *supra* note 9.

[66] Kuppler et al., *supra* note 49 at 15.

[67] Kasy and Abebe, *supra* note 6; Binns, *supra* note 63; Cooper, Abrams, and Na, *supra* note 9; Alejandro Noriega-Campero et al., *Active fairness in algorithmic decision making*, in PROCEEDINGS OF THE 2019 AAAI/ACM CONFERENCE ON AI, ETHICS, AND SOCIETY 77 (2019); Sanghamitra Dutta et al., *Is there a trade-off between fairness and accuracy? a perspective using mismatched hypothesis testing*, in INTERNATIONAL CONFERENCE ON MACHINE LEARNING 2803 (2020); Irene Chen, Fredrik D. Johansson & David Sontag, *Why is my classifier discriminatory?*, 31 ADVANCES IN NEURAL INFORMATION PROCESSING SYSTEMS (2018); Michiel A. Bakker et al., *On fairness in budget-constrained decision making*, in PROCEEDINGS OF THE KDD WORKSHOP ON EXPLAINABLE ARTIFICIAL INTELLIGENCE (2019); Hu and Chen, *supra* note 57.

[68] Cooper, Abrams, and Na, *supra* note 9; Hilde Weerts, Lambèr Royakkers & Mykola Pechenizkiy, *Does the End Justify the Means? On the Moral Justification of Fairness-Aware Machine Learning*, (2022), http://arxiv.org/abs/2202.08536 (last visited Jul 13, 2022).

[69] Kuppler et al., *supra* note 49.

The validity of the objection and its ability to "defeat" (strict) egalitarianism is the subject of significant debate in philosophy.[74] In practice it is often avoided by attributing it to a misreading of the principle of equality,[75] or by noting that strict egalitarians are a rare breed and most "sensible egalitarians" are pluralists,[76] meaning they value other goods and do not measure justice solely in terms of equality.[77][78] At most, then, it would appear levelling down would only be accepted as inherently good by strict egalitarians. Levelling down is rejected as an acceptable solution to distribution problems by other schools of thought for whom justice is not simply a matter of equality but instead requires consideration of welfare, utility, priority, luck, and similar properties.[79]

---

[70] Campbell Brown, *Giving up Levelling Down*, 19 ECONOMICS & PHILOSOPHY 111 (2003).

[71] Holtug, *supra* note 11.

[72] *Id.*

[73] As an extreme example of the objection, Christiano and Braynen offer the example of a lifeboat with three passengers which will sink unless one passenger is thrown overboard. The principle of equality suggests that equal welfare can only be achieved by levelling down, meaning nobody is thrown overboard and all the passengers die. This egalitarian outcome would be preferable to an inegalitarian outcome in which one passenger is sacrificed so that the remaining two passengers can survive or have higher welfare. This course of action clearly conflicts with a common sense understanding of justice. See: Christiano and Braynen, *supra* note 59.

[74] Brown, *supra* note 70; Christiano and Braynen, *supra* note 59; Holtug, *supra* note 11; Doran, *supra* note 11; Michael Otsuka & Alex Voorhoeve, *Equality versus priority*, *in* OXFORD HANDBOOK OF DISTRIBUTIVE JUSTICE 65 (Serena Olsaretti ed., 2018); Larry S. Temkin, *EQUALITY, PRIORITY OR WHAT?*, 19 ECONOMICS & PHILOSOPHY 61 (2003); Richard J. Arneson, *Egalitarianism and Responsibility*, 3 THE JOURNAL OF ETHICS 225 (1999); Harry Frankfurt, *Equality as a Moral Ideal*, 98 ETHICS 21 (1987).

[75] Brown, *supra* note 70; Christiano and Braynen, *supra* note 59.

[76] Otsuka and Voorhoeve, *supra* note 74; Parfit, *supra* note 11 at 85.

[77] Brown, *supra* note 70.

[78] Brown, for example, has suggested that advocates of the levelling down objection must explain "what it means, in their view, to say that one thing is better than another in a respect" for levelling down to be a valid objection to egalitarianism. See: *Id.* at 113. One example of a pluralist reconstruction that combines equality and welfare to defeat the objection is the "*common good conception* of the principle of equality" which "favours

The success of the objection is ultimately irrelevant for our purposes, but it helpfully draws focus to a key question for the future of fairML: under what theoretical or practical conditions, if any, can levelling down to enforce group fairness be justified?

## 4.1   The value of equality

Egalitarians believe equality has value. This deceptively simple statement hides significant theoretical complexity.[80] Egalitarian approaches can be distinguished according to the type of value they assign to equality. For strict egalitarians, equality has *intrinsic value,* meaning equality is good in itself. Likewise, inequality is bad in itself, even when it has no bad effects.[81] The intrinsic value of equality is reflected in formal notions of equality and the principle of equal treatment according to which similar individuals must be treated

---

states in which everyone is better off to those in which everyone is worse off." See: Christiano and Braynen, *supra* note 59 at 395. Pluralist egalitarians "believe that it would be better both if there was more equality, and if there was more utility." See: Parfit, *supra* note 11 at 85. Both equality and utility are thus given moral weight by pluralists.

[79] Pluralist egalitarian approaches, for example, recognise that there are other goods besides equality that should be considered in assessing distributive justice. These alternative goods can explain why it is better to achieve equality by making people better off than worse off and thus avoid the levelling down objection. Similarly, some egalitarians argue that strict egalitarianism is not required to achieve equality of treatment or opportunity. See: Binns, *supra* note 7. Rather, a maximin distribution is sufficient according to which inequalities are tolerated so long as they benefit disadvantaged groups. See: John E. Roemer, *Equality of opportunity: A progress report*, 19 SOCIAL CHOICE AND WELFARE 455 (2002); JOHN RAWLS, A THEORY OF JUSTICE, REVISED EDITION (1999). Prioritarianism is concerned principally with absolute entitlements rather than comparative levels of welfare Parfit, *supra* note 11.. Distribution principles which maximize weighted utility across groups are preferred, with priority given to benefits to the worse off, and inequalities between groups found acceptable if they lead to greater utility. See: Otsuka and Voorhoeve, *supra* note 74; Parfit, *supra* note 11; Parfit, *supra* note 52. Approaches to prioritarianism differ on how they measure priority and the size of benefits across groups. See: LIPPERT-RASMUSSEN, *supra* note 53. A maximin approach, for example, focuses solely on benefits to the worst off. See: RAWLS; Gerald A. Cohen, *On the currency of egalitarian justice*, 99 ETHICS 906 (1989). Relatedly, sufficientarianism favours distribution principles which ensure all people receive at least a minimum threshold of resources to ensure a good quality of life. Inequalities are tolerated once this minimum level is met across relevant groups. See: Frankfurt, *supra* note 74; Jonathan Herington, *Measuring Fairness in an Unfair World*, in PROCEEDINGS OF THE AAAI/ACM CONFERENCE ON AI, ETHICS, AND SOCIETY 286 (2020), https://dl.acm.org/doi/10.1145/3375627.3375854 (last visited Jul 13, 2022).

[80] Amartya Sen, *Equality of what?* (1980).

[81] TEMKIN, *supra* note 11; Parfit, *supra* note 11.

similarly regardless of their ethnicity, sex, gender, and other protected characteristics.[82] Prohibitions against direct discrimination or disparate treatment, for example, prohibit "less favorable" treatment on these grounds.

If equality has intrinsic value, a situation is measurably improved if distributions are equal between groups, regardless of the effect of achieving parity on the groups in question. The harm of inequality arises from the mere fact that some group of people are worse off than others.[83] For example, achieving parity in the distribution of public housing by eliminating public housing altogether would be inherently good for a strict egalitarian in at least one sense, independent of the severe impact of such a policy on tenant welfare, because inequality had been removed.[84] However, some strict egalitarian theorists attach further conditions to the inequality to be 'bad' or undeserved, for example, it must have arisen through no fault or choice of the individual.[85]

The intuitive problem with formal equality is that equal treatment is a comparative measure and need not be concerned with absolute levels of welfare or benefit. Equal treatment can be achieved "whether the two individuals are treated equally well or equally badly," or by "removing a benefit from the relatively privileged group."[86] Intrinsic value alone cannot explain the intuition that it is better to achieve equality by bringing all relevant groups up to an equal level as opposed to making them equal but worse off. For that, appeal to some instrumental value of equality is necessary.[87] Equality can be said to have *instrumental value* derived from the good effects it produces. It is good as a means to achieve some other valuable goal related to justice,[88] such as universal freedom, the development of human capacities and personality, mitigation of suffering and stigmatisation, or avoiding conflict or envy between people created by inequality.[89]

Inequality may be instrumentally bad, for example, because of the social injustice it entails.[90] There are two types of injustice to consider: (1) a comparative sense of justice, meaning we are concerned with the mere fact that people are treated differently from others, or do not receive their fair share of a resource; or (2) a non-comparative sense of justice, where injustice arises because a person is not treated as they deserve, independently of any consideration of how others are treated. In terms of levelling down, a comparative sense of justice alone would not reject eliminating disparity by treating everyone neutrally but equally unjustly,[91][92] whereas a non-comparative sense would reject it on the basis that people are treated unjustly in absolute terms, for example being denied a share of a vital resource.[93]

---

[82] Wachter, Mittelstadt, and Russell, *supra* note 10; Sandra Fredman, *Substantive equality revisited*, 14 IJCLAW 712 (2016).

[83] Parfit, *supra* note 11.

[84] *Id.* at 98.

[85] Parfit, *supra* note 11; Cohen, *supra* note 79; TEMKIN, *supra* note 11; Richard J. Arneson, *Equality and equal opportunity for welfare*, 56 PHILOSOPHICAL STUDIES 77 (1989); THOMAS NAGEL, EQUALITY AND PARTIALITY (1995).

[86] Fredman, *supra* note 82 at 717–8.

[87] Otsuka and Voorhoeve, *supra* note 74.

[88] TEMKIN, *supra* note 11; Parfit, *supra* note 11.

[89] THOMAS M. SCANLON, THE DIVERSITY OF OBJECTIONS TO INEQUALITY (1996); THOMAS SCANLON, WHY DOES INEQUALITY MATTER? (2018).

[90] Parfit, *supra* note 11 at 88.

### 4.1.1 Performance, utility, and harms

Unless one is a strict egalitarian believing in the intrinsic value of equality above all else, levelling down can only be justified by appeal to some instrumental value of greater equality between groups that offsets the harm it causes through reduction of performance or access to a valuable good. It is intuitively difficult to claim that equality has instrumental value in a case where no group experiences a direct benefit. This draws on a key intuition in moral philosophy called the person-affecting view,[94] which says "that the part of morality that is concerned with outcomes should take a person-affecting form, i.e. that what makes good outcomes good and bad is how they *affect people*… it is because levelling down affects no-one for the better that it cannot seem to make an outcome better."[95]

The harm of levelling down can thus be linked to the inherent value of the good or benefit being distributed (e.g., performance, recall). Indeed, this is the core observation behind alternative theories to strict egalitarianism such as pluralist egalitarianism, prioritarianism, and welfarism: justice is not derived solely from equality but must also account for the utility of distributed goods and their impact on recipients' welfare, rights, and other interests. Following from our motivating example of classifiers used in cancer screening (see: Section 2.2), a reduction in classifier accuracy for any group constitutes a welfare harm by creating more cases of misdiagnosis for that group. In other words, performance is inherently valuable for patient health and welfare. It may, of course, be possible to justify this welfare harm if it improves performance for another worse performing or otherwise

---

[91] LIPPERT-RASMUSSEN, *supra* note 53.

[92] Discrimination, like equality, is essentially comparative—it must be possible to show that someone is better or worse off than someone else to say discrimination has occurred. See: *Id.*

[93] Parfit, *supra* note 11.

[94] TEMKIN, *supra* note 11; Holtug, *supra* note 11; Doran, *supra* note 11; Parfit, *supra* note 11.

[95] Holtug, *supra* note 11 at 167.

priority group, but the existence of the harm to the welfare of the advantaged group remains constant.[96]

Of course, equal predictive accuracy does not guarantee equal outcomes. Imagine two groups who have equal predictive accuracy but significantly different false positive and false negative rates. In the context of healthcare this can equate to one group being much more frequently misdiagnosed as healthy and the other misdiagnosed as sick, meaning each group receives undertreatment or overtreatment.[97] Depending on the context each option may be seen as preferable or problematic, as "different types of errors have different costs."[98] Overtreatment, for example, has been linked to significant harms from unnecessary treatments including elective surgery in the context of mammography screening.[99] However, in the case of Rabies diagnosis, even if a diagnosis comes back negative, the cost of being wrong necessitates proceeding with treatment of Rabies.[100] The value of the distributed goods or benefits and the cost of different types of errors thus have a direct influence on the acceptability of levelling down in practice.[101]

Reflecting the inherent value of distributed goods, claimants in equality law cases typically do not seek a remedy that involves levelling down; rather, they also want to be included in some benefit that is currently denied to them, and implicitly propose an "alternative distributive principle" which would, in their estimation, more equitably distribute the benefit. As Reaume explains, "…levelling down is rarely the remedy litigants pursue: they ask to be allowed to vote as well, not that voting be abolished, or that a pension scheme include them, not that it be repealed," as doing so "would deprive everyone of something

---

[96] The idea that performance can be inherently valuable and loss of performance harmful, while seemingly obvious, is not universally acknowledged in fairML. In a paper endorsing strict egalitarianism for fairness measures, for example, Kuppler et al. observed that "there is no obvious reason why some individuals should deserve or need a higher probability of prediction errors than others." See: Kuppler et al., *supra* note 49 at 12. Clearly this approach ignores the real harms caused by lowering performance for advantaged groups (e.g., missing more cases of cancer; see: Section 2.2) in the search for perfect equality in error rates.

[97] Deborah Hellman, *MEASURING ALGORITHMIC FAIRNESS*, 106 VIRGINIA LAW REVIEW 56 (2020).

[98] *Id.* at 828–9.

[99] PETER C. GØTZSCHE, MAMMOGRAPHY SCREENING: TRUTH, LIES AND CONTROVERSY (1 ed. 2012).

[100] Hellman, *supra* note 97.

[101] *Id.*; Michael Pace, *The Epistemic Value of Moral Considerations: Justification, Moral Encroachment, and James' "Will To Believe,"* 45 NOÛS 239 (2011); Brent Mittelstadt et al., *The ethics of algorithms: Mapping the debate*, 3 BIG DATA & SOCIETY (2016), http://bds.sagepub.com/lookup/doi/10.1177/2053951716679679 (last visited Dec 15, 2016).

all are properly entitled to, and thus exacerbate rather than solve the problem."[102]

Levelling down is harmful because it denies access to a valuable resource to more people than is strictly necessary. Something of value is lost or removed from an advantaged group to reduce disparity. This provides no direct improvement in utility (e.g., performance, welfare) to disadvantaged groups. There is no one for whom a levelled down situation is measurably improved, and it will be actively harmful in cases where performance is inherently valuable. Its instrumental value can only be measured indirectly as benefits to future distribution scenarios or opportunities because whatever utility is lost (e.g., goods, opportunities, access to resources) is not re-distributed to disadvantaged groups (if that was the case, we would not be speaking of an instance of levelling down).

For now, we can conclude that levelling down can be rejected on the basis that it produces no instrumentally valuable direct benefits to disadvantaged groups (e.g., improved utility, welfare, priority). It may, nonetheless, be justifiable if (1) we solely value equality above all else as strict egalitarians or (2) can appeal to the instrumental value of its indirect effects. We have already considered the first option and will now turn to the second in examining the substantive equality harms and opportunities created by levelling down.

## 4.2   Substantive equality harms

There are a range of possible harms of inequality and levelling down which cannot be measured in terms of strict equality or solved through equal treatment. Notions of substantive equality draw attention to the subtle qualitative harms of levelling down which are not captured in simple utilitarian calculations focused on "the well-being of the advantaged group and the costs of relinquishing inequality's privileges."[103] Focusing solely on relative disadvantages between groups can ignore "stigma, stereotyping, humiliation, and violence on grounds of gender, race, disability, sexual orientation, or other status" experienced by disadvantaged individuals.[104] Relational and identity-based harms arising from misrecognition, denigration or humiliation are likewise not captured by utility. As Brake summarises, the ability to "participate on equal terms in community and society more

---

generally" is fundamentally valuable but not fully captured by strict or formal equality.[105]

Ideal solutions to inequality should fully remedy "all of the injuries, material and nonmaterial, to the persons disadvantaged by inequality," and not simply deal in redistribution of goods or resources.[106] Levelling down fails to address the social and relational harms of inequality. As explained by Fredman:

> "Where the injury inheres in the materially dissimilar treatment of persons otherwise similarly situated, it may be remedied by eliminating the differential treatment either by leveling down, leveling up, or setting a baseline at some point in between. Where the injuries from discrimination transcend the material consequences of differential treatment and are social or relational in nature, however, leveling down may exacerbate the injuries of discrimination and is not consistent with equality law."[107]

Restricting access, lowering performance, removing goods, or otherwise levelling down rather than levelling up expresses "unequal concern" and can solidify pre-existing social inequality and cause stigmatisation, social backlash, and undermine solidarity between groups in society.[108] Historically advantaged groups may indeed prefer to surrender a benefit rather than extend access going forward as a way to maintain their relative privileged position and solidify "status hierarchies."[109]

Social and relational harms of levelling down can also be captured in pluralist egalitarian approaches. Social and political egalitarians believe that "material and social inequalities are bad when and because they undermine individuals' ability to live as equal citizens who are willing to offer and abide by fair terms of social cooperation."[110] Equality in this context has instrumental value because inequalities can divide communities by giving rise to "morally problematic attitudes," such as "servility, envy, and a lack of self-respect among the worst off and arrogance and a jealous guarding of relative advantage among the better off."[111]

---

[105] *Id.* at 732.

[106] Brake, *supra* note 103 at 539.

[107] *Id.* at 560.

[108] *Id.* at 607.

[109] *Id.* at 578.

[110] Otsuka and Voorhoeve, *supra* note 74 at 65; Richard Norman, *The social basis of equality*, 10 RATIO 238 (1997); Elizabeth S. Anderson, *What Is the Point of Equality?*, 109 ETHICS 287 (1999); Martin O'neill, *What Should Egalitarians Believe?*, 36 PHILOSOPHY & PUBLIC AFFAIRS 119 (2008).

[111] Otsuka and Voorhoeve, *supra* note 74.

Examples of stigma and relational harms abound in equality law jurisprudence. In the USA, *Cazares v. Barber[112]* dealt with a student who was denied entrance to the National Honor Society (NHS) on the basis of pregnancy, marriage status, and living situation. The student won her case against the school district on Title IX and Fifth Amendment grounds. In response, the school district cancelled the ceremony and ended their participation in the NHS programme. The student did not gain access to the benefit sought and caused the benefit of the programme to be withdrawn from all students. Another example comes from *Heckler v. Mathews* (465 U.S. 728 1984) in which male plaintiffs sought access to spousal benefits under the Social Security Act which were available to "wives and widows, but not husbands or widowers." The court chose not to extend the benefits on the basis that "the injury in an equality claim inheres in the stigma from the discriminatory treatment and not the deprivation of the material benefit itself," meaning the injury could be remedied by enforcing equal treatment "denying benefits to women rather than extending them to men."[113] Both cases were remedied by enforcing equal treatment through levelling down, and thus do not reflect unequal concern for the affected groups. Nonetheless, both solutions can be characterised as causing relational and stigma-based harms to the claimant in their self-perception and relationship with others in their communities.[114]

Other classic examples of levelling down which undermine social solidarity come from the era of racial desegregation in the United States. In *Palmer v. Thompson* (403 U.S. 217 220-21 1971) the city of Jackson, Mississippi responded to an order to desegregate its public swimming pools by closing all of them rather than integrating. In this case, the city levelled down by removing pool access for the advantaged group (i.e., white citizens) rather than extending it to the disadvantaged group (i.e., Black citizens). The decision was upheld by the Court on the basis that all citizens do not have a 'right to a pool', or an affirmative duty for the city to operate a pool. Similarly, in *Griffin v. Cty. Sch. Bd. Of Prince Edward City* (377 U.S. 218 1964), in response to a desegregation order, the school district closed public schools and opened 'private' schools in their place that served only white students. The decision was eventually found to constitute an equal protection violation, not because the public schools were closed, but rather because the 'private' schools continued to receive support from the state and county.[115]

---

[112] Cazares v. Barber, No. CIV-90-0128-TUC-ACM (D. Ariz. May 31,1990), afftd, 959 F.2d 753 (9th Cir. 1992).

[113] Brake, *supra* note 103 at 593.

[114] Brake, *supra* note 103.

[115] Thomas B Nachbar, *Algorithmic fairness, algorithmic discrimination*, 48 51, 554 (2020).

Levelling down to enforce equal treatment is not a uniquely American solution to inequality.[116] In *A. v. Secretary of State for the Home Department* (2004 U.K.H.L. 56 (HL)) a challenge was posed to legislation that granted "authorities the power to detain non-UK nationals indefinitely without trial if they were suspected of international terrorism." The legislation was struck down by the House of Lords on the basis that non-UK and UK nationals were "alike" in their capacity to commit terrorism and should be treated alike. In response, the government levelled down by extending their power for indefinite detention to both UK and non-UK nationals, ensuring equal treatment by "intruding equally on the liberty of both groups."[117]

While aligned with formal equality, the solutions in these cases can still be criticised on substantive equality grounds. Strictly speaking, levelling down in these cases treated all groups "the same in material respects." But this focus solely on equal treatment misses how such solutions "express selective disdain or disregard for some persons," and reproduce or reinforce inequality through the "expressive meaning" of the judgement or corrective action.[118] One example where courts seemingly recognise the significance of expressing meaning is found in *Johnson v. California* (543 U.S. 499 2005), which dealt with racial segregation of prisoners. The U.S. Supreme Court rejected the idea that racial segregation can be considered neutral or acceptable if all racial groups are segregated to an equal degree.[119] This judgement follows the spirit of the historic *Brown v. Board of Education* decision, in which the Court found that "the state's segregation…expressed a message of racial inferiority.."[120]

Somewhat surprisingly, levelling down has been repeatedly upheld as a legitimate solution for legal frameworks prioritising formal equality, such as US equality law.[121][122] The same does not hold true for legal frameworks prioritising substantive equality, such as EU non-discrimination law, where the permissibility of levelling down is much more complex [123] (see: Section 4.2.1). Nonetheless, the preceding examples show that people seeking remedies under equality law are often left materially equal to others, but substantively worse off in terms of stigmatisation and social solidarity. Equal treatment through levelling down excludes disadvantaged groups from the valuable benefit sought and expresses a preference by advantaged groups for "losing the benefit rather than broadening the community of persons sharing in it." Levelling down solidifies "social stratification" by ensuring the "separateness and social inequality" between advantaged and disadvantaged people remains unchanged.[124] This trend is intuitively concerning because it conflicts with the intuition that "a person who is harmed by discrimination and successfully prosecutes a discrimination claim should benefit from the suit and that persons should not be made worse off unnecessarily."[125]

---

[116] It is worth noting that levelling down is not a historic relic of equality law. According to Brake, "the underlying premise-that equality law has little or nothing to say about leveling down as a response to inequality-has remained largely unchallenged" between the 1960s and early 2000s. See: Brake, *supra* note 103 at 519–20.

[117] Fredman, *supra* note 82 at 717–8.

[118] Brake, *supra* note 103 at 571.

[119] Fredman, *supra* note 82 at 724.

[120] Brake, *supra* note 103 at 572–3.

[121] Fredman, *supra* note 82; Brake, *supra* note 103.

[122] It is worth noting that levelling down may be more readily accepted under US equality law because of the predominant focus on formal equality. In jurisdictions that favour a substantive approach to equality, which focuses on equality of opportunity rather than equality of treatment, levelling down is more problematic. See: Wachter, Mittelstadt, and Russell, *supra* note 10.

### 4.2.1   Levelling down for social change

The legal permissibility of levelling down suggests an impoverished conceptualisation of equality may lie at the basis of current legal frameworks in the jurisdictions discussed due to their failure to redress relational and stigma-based harms of materially equal treatment.[126] However, this conclusion is not yet merited, as levelling down can also be used to pursue valuable second-order or indirect substantive equality benefits that are instrumentally valuable for (historically) disadvantaged groups. These include (1) removing unjustified entrenched advantages or (2) for civil action.

Considering the first, levelling down may be justifiable if used as a mechanism to eliminate an unjustified entrenched advantage. Unequal treatment in favour of the disadvantaged may be necessary in cases where equal treatment would further solidify or exacerbate an "antecedent disadvantage."[127] We refer to this as the "levelling the playing field" aim of substantive equality. It applies in cases where removing an antecedent advantage is necessary to ensure equal access to rivalrous or scarce resources, or in cases where an entrenched social or institutional advantage prevents equal consideration, capabilities, or opportunities in the future.[128] Equality here can have instrumental

---

[123] *Id.*; Fredman, *supra* note 82; TARUNABH KHAITAN, A THEORY OF DISCRIMINATION LAW (2015); Sophia Moreau, *What Is Discrimination?*, PHILOSOPHY & PUBLIC AFFAIRS 143 (2010); DEBORAH HELLMAN & SOPHIA MOREAU, PHILOSOPHICAL FOUNDATIONS OF DISCRIMINATION LAW (2013); Wachter, Mittelstadt, and Russell, *supra* note 26.

[124] Brake, *supra* note 103 at 575.

[125] *Id.* at 540.

[126] Wachter, Mittelstadt, and Russell, *supra* note 10.

[127] Fredman, *supra* note 82 at 718.

[128] Fredman, *supra* note 82; AMARTYA SEN, INEQUALITY REEXAMINED (1995).

social value by improving opportunities for relevant connected groups in a community.[129]

Consider, for example, access to employment or education. Direct material weakening of the competitiveness of advantaged groups, for example barring men from university degree programmes, runs counter to the substantive aims of equality law. Levelling the playing field does not necessarily disrupt entitlements of advantaged groups; rather, it aims to remove pre-existing exclusionary standards or arbitrary barriers to equal access or opportunities from the relevant decision-making process or distribution principle (e.g., college admissions requiring a degree from male only schools).[130] Here, levelling down is justified because the equality injury experienced by advantaged groups is necessary to realise benefits for disadvantaged groups.[131]

Ideally, corrective equality actions should not only improve equality in terms of results, access, capabilities, or opportunities, but also account for the social and institutional structures responsible for the inequality or entrenched advantages in question.[132] Take for example gender equality in college athletics funding where high prestige male athletics programmes have historically been given much more funding than equivalent female programmes or lower prestige male programmes. Extending current funding levels to other programmes would be unsustainable for most colleges. In such cases "equality law should permit some leveling down to find a baseline that is not based on male privilege."[133]

Levelling the playing field is typically inappropriate in cases dealing with fundamental rights or goods, or non-rivalrous goods with inherent value such as recall or accuracy in cancer screening (see: Section 2.2). Extending the right to vote to women, for example, could not have been achieved by denying the right to men, and would have not been consistent with considerations of liberty. Similarly, reducing recall for male patients increases undiagnosed cases of cancer and is not strictly necessary to improve recall for female patients (see: Section 6).

Considering the second, levelling down can also be used as a type of civil action to force reconsideration of problematic social and institutional norms that contribute to unequal concern. A standout example of this justification came in the extension of marriage rights to same-sex couples in Benton County, Oregon. In response to a lawsuit filed to block the extension, the county levelled down by suspending all marriage licenses until the validity of the state's marriage law was resolved. As Brake explains:

> "…the leveling down occurred not as resistance to the equality challenge by gays and lesbians to the state's marriage laws, but in furtherance of it…the leveling down decision was not a defensive construction of social meaning designed to reinforce a status hierarchy disparaging gay and lesbian couples. Instead, it was a tactic designed to challenge that status hierarchy and hasten the extension of marriage to gay and lesbian couples by equalizing the status of their relationships. Importantly, the measure was understood by members of the gay and lesbian community as breaking down, rather than reinforcing, status differentials between gay and straight couples. Finally, it is significant that the leveling down of marriage was designed as a temporary measure, as part of a larger strategy to ultimately extend the privilege of marriage to same-sex couples."[134]

---

[129] Parfit, *supra* note 52; Parfit, *supra* note 11.

[130] Wachter, Mittelstadt, and Russell, *supra* note 10; Brake, *supra* note 103; Kim, *supra* note 30; Wachter, Mittelstadt, and Russell, *supra* note 26.

[131] Another example is land ownership, which has historically been limited to certain genders or royalty; extending access would require removing this privilege from historically advantaged groups.

[132] Fredman, *supra* note 82.

[133] Brake, *supra* note 103 at 594–5.

This step was viewed as a positive civil action in support of the LGBTQ+ community's push for marriage equality at a state level. A temporary solution of levelling down across all groups by suspending all marriage licenses was used to ensure equal concern rather than mere equal treatment.[135]

In fairML, levelling down for civil action could similarly be used as means to force consideration of fairness in a production environment. Researchers and developers can use levelling down to delay or prevent deployment of models with unjustifiably poor performance for disadvantaged groups. By lowering performance for an advantaged group, a developer could prevent deployers from having access to a 'high accuracy' or 'unbiased' model, and instead force levelling up by design until the model performs acceptably well across all groups (see: Section 6). Temporarily reducing performance can economically impact the deployer in cases where the advantaged group is also the largest (potential) customer base, or the group with the most socioeconomic or political power. As a result, levelling down can empower disadvantaged groups by making highly disparate models less commercially viable.

---

[134] *Id.* at 600. Brake offers another illustrative example drawn from college athletics. Male athletes are afforded certain privileges based on a problematic notion of masculinity which Brake argues would not be appropriate to extend to female athletes. "At the prestigious level of NCAA Division I-A football, for example, it is a common practice to have the football team housed in a hotel the night before home games. The rationale typically rests on the difficulty of otherwise controlling and disciplining the players to avoid the kind of behavior that would hurt their game performance. The practice is based on a model of a male athlete who embodies a ruggedly uncontrollable masculinity and it is applied uniquely to football players. Extending such a practice to female athletes, at least on the same rationale, would make little sense. Instead, equality should require readjusting the athletic model upon which the practice is based to a gender-inclusive standard that holds all athletes responsible for their own behavior." See: *Id.* at 597–8.

[135] Brake, *supra* note 103 at 561; RONALD DWORKIN, TAKING RIGHTS SERIOUSLY (2013).

All of these possible justifications to level down for social change are of course highly contextual and dependent on the circumstances of specific distribution problems and policies. Here we intended only to outline possible justifications and the considerations which may be offered in their favour, not to conclusively say whether a particular aim constitutes a justified basis for levelling down in all cases. Applying such justifications to cases of algorithmic fairness faces additional difficulties, to which we now turn.

## 5    JUSTIFYING LEVELLING DOWN IN FAIRML

There are many reasons to reject levelling down in fairML: (1) it unnecessarily and arbitrarily harms advantaged groups in cases where performance is intrinsically valuable; (2) it demonstrates a lack of equal concern for affected groups and can undermine social solidarity and contribute to stigmatisation; (3) it fails to live up to the substantive aims of equality law and fairML, and squanders the opportunity afforded by interest in algorithmic fairness to substantively address longstanding social inequalities; and (4) it fails to meet the aims of many theories of distributive justice including pluralist egalitarian approaches, prioritarianism, sufficientarianism, and others. But the question remains: when, if ever, can levelling down be justified in fairML?

If one is a strict egalitarian concerned only with equal treatment or is enforcing group fairness to satisfy a social policy or law that requires strict egalitarianism, levelling down in fairML can be justified. Beyond these straightforward situations, the arguments discussed above which could justify levelling down in practice are a poor fit for fairML (see: Section 4). Possible justifications address contextual factors such as available resources, current distributions, historical inequalities, and their impacts, and aim to achieve goals such as civil action or removing or forcing questioning of an unjustified pre-existing advantage. They turn on the realisation of an underlying goal that is independently justified within a given legal, ethical, political, or social framework (e.g., questioning heteronormativity of marriage, removing entrenched advantages). In such cases, levelling down is an imperfect means to realise some agreed upon end of greater value, meaning the harm of eliminating utility for the sake of parity has clear instrumental value.[136]

The problem is that enforcement methods and practices in fairML currently do not engage with possible justifications and criticisms of the distribution principles they produce.[137] Simply put, fairness is treated as a standardized mathematical problem to be solved. Justifying how a measure is satisfied in practice, linking it to some underlying equality goal, and exploring whether a less equal but less harmful path would be preferable are rarely part of enforcing fairness.[138] Debates in distributive justice recognise that "different people may value the same outcome or set of harms and benefits differently." This fact is not reflected in the tendency in fairML to assume "a uniform valuation of decision outcomes across different populations" [139] and use cases, which reduces a highly complex, value-laden debate and set of theories and decisions to an oversimplified homogenous set of assumptions.

---

[136] Strict egalitarians may disagree with this assessment, citing the inherent value of parity, but this view is controversial in law and philosophy (see: Section 4).

The same type of error can cause substantially different types of harm depending on the use case. Take facial recognition as an example ML application. If facial recognition is used by police to identify people with outstanding warrants in crowds, the harm of a false positive is an unjustified arrest. In contrast, if it is used to track perceived compliance with visa requirements,[140] the harm of false negatives is perceived non-compliance with a monitoring regime that could have significant legal ramifications (e.g., deportation). The harms of false positives and negatives likewise vary for facial recognition used for loan decisions or job interviews. If the enforcement of fairness in ML is to resemble comparable legal decisions (where, as we have seen, levelling down can be justified), it is essential to consider the specific types and severity of harms actually suffered by affected populations.

Researchers, developers, and deployers of 'fair' ML systems do not currently seriously engage with such questions at a local level. Levelling down is arguably not viewed as something that requires justification. At most, one need only justify the choice of fairness measure; the steps taken to satisfy it in practice are normatively irrelevant.[141] At best, tenuous connections are drawn between fairness measures and complementary political or ethical theories.[142][143] Works that merely link methods and measures to complementary theories of equality suggest that researchers using those methods and measures believe in those theories, or have explicitly chosen them, and have

---

critically thought about the theory of equality their models should promote.[144][145] This cannot be taken for granted. Rather, the vast majority of cases of levelling down are unintentional and invisible, resulting from convenience and the limited ways performance and (dis)parity are currently measured ,[146] not theoretical conviction or justification.

It is likewise unclear whether substantive equality goals can be achieved directly through enforcement of fairness measures on ML models. Intentional levelling down to create a worse performing classifier, for example, can draw attention to a problematic performance gap in order to prompt civil action. Levelling down does not produce this effect by itself. Rigid enforcement of fairness measures does not allow for external corrective action to reduce harm because measures must be solvable with the data at hand.[147] In cases of competition over limited resources, a bigger 'piece of the pie' requires a smaller piece to be given to someone else. FairML typically cannot make this sort of trade-off explicitly because models do not have a picture of 'how big the pie is', or awareness of the limitations of the resources at hand.[148] Nonetheless, actions which are not directly quantifiable or within the control of the modeller, such as collecting more data on equality-relevant features (e.g., socioeconomic status, prior opportunities) or increasing available resources (e.g., cancer screening) in the production environment, are typically not considered but could be viable means to avoid levelling down in practice.[149]

---

[144] Binns, *supra* note 7.

[145] We have made a similar observation in prior work introducing the notion of bias preservation, where we argued that researchers, developers, and deployers of ML systems need to explicitly choose the biases their models should exhibit. See: Wachter, Mittelstadt, and Russell, *supra* note 10. Our argument here is complementary but distinct; we argue that people working in fair ML need to be more explicit and reflective about the underlying goals their choice of fairness measures and methods supports.

[146] Kuppler et al., *supra* note 49.

[147] Binns, *supra* note 63.

[148] Enforcing fairness typically involves balancing output rates (e.g., acceptance rates, sufficiently high recall for cancer detection) between groups. For specific instances of levelling down in fair ML to be justified under something like the 'levelling the playing field' argument (see: Section 4.2.1), models would need to be distributing a known limited quantity of a set of outputs. In practice, this is not how classifiers operate. This observation does not, however, preclude justification of levelling down at a general level. Decision-makers may, for example, choose to lower performance for specific historically advantaged groups for all classifiers used in a given sector or for classifiers considering specific historically disadvantaged groups in order to level the playing field.

[149] Binns, *supra* note 63; Cooper, Abrams, and Na, *supra* note 9. Collection of data on equality-relevant features is not the same as collecting more representative data to combat bias against data impoverished groups, which is a common approach in the field and can avoid the need for levelling down. This is sometimes called 'active fairness'. See: *Id.* This approach helps mitigate biases in the existing data affecting disadvantaged groups without directly impacting advantaged group performance or outcomes. For examples see: Obermeyer and Mullainathan, *supra* note 20; Noriega-

To bring fairness in machine learning out of a testing and research environment and into systems making important real-world decisions,[150] theoretically richer and contextually sensitive work cannot be the only answer. Instead of simply levelling down out of convenience to 'solve' fairness and arbitrarily harming people in the process, fairML must shift to a harms-based framing and 'level up' systems by design.

## 6   LEVELLING UP BY DESIGN WITH MINIMUM RATE CONSTRAINTS

Levelling up can be understood as a new type of constraint for fairness that serves as an alternative to strict egalitarianism achieved through levelling down. If we believe that particular groups are harmed by decision rates or recall that is too low, we can simply increase them to the required level. Instead of requiring that harms are equalized between groups, levelling up instead requires that harms are reduced to, at most, a given level per group. For example, if we believe that people are being harmed by low selection rates, precision, or recall, instead of enforcing that these properties be equalised across groups, we can instead require that every group has, at least, a minimal selection rate, precision, or recall. We refer to this type of minimum acceptable threshold for harms visited upon groups in the pursuit of fairness as a "minimum rate constraint" (MRC).

Unlike existing methods that enforce strict equality, there is no direct gain from decreasing the rate for any group when fairness is defined in terms of minimum rate constraints. Rather, the focus of fairness is on levelling all groups up to a minimally acceptable performance threshold. Performance reductions are only tolerated if they are causally necessary to improve the situation of another group.

We show how levelling up can be achieved through MRCs in practice by using post-processing. The family of post-processing methods we consider [151] tune a separate offset for each group, which alters the proportion of individuals in each group that receive a positive decision. By altering these thresholds, it is possible to enforce (either approximately or exactly) a wide range of fairness measures. Through changes to these thresholds we obtain a set of classifiers with varying accuracy and fairness. By discarding all bad classifiers that are both

---

Campero et al., *supra* note 67; Dutta et al., *supra* note 67; Chen, Johansson, and Sontag, *supra* note 67; Bakker et al., *supra* note 67; Lucas Dixon et al., *Measuring and mitigating unintended bias in text classification*, in Proceedings of the 2018 AAAI/ACM Conference on AI, Ethics, and Society 67 (2018); Ruchir Puri, *Mitigating bias in AI models*, IBM Research Blog (2018); Heidi Ledford, *Millions of black people affected by racial bias in health-care algorithms*, 574 Nature 608 (2019).

[150] Bakalar et al., *supra* note 25.

[151] Corbett-Davies et al., *supra* note 15.

less accurate and less fair than at least one other classifier in the set, we obtain a *Pareto Frontier* that consists of classifiers with the best possible fairness and accuracy trade-offs.

We will work through two examples where we evaluate fairness against accuracy on the Adult Dataset to show how levelling down results from enforcing fairness measures as currently conceived, and how it can be avoided through a shift in focus from parity to minimum thresholds, or what we refer to as 'minimum rate constraints'. First, we will use the standard equality notion of demographic parity as our fairness measure of choice and show how it induces levelling down. We will then compare this result to a second option, where we trade off the minimum selection rate for each group against accuracy. This approach induces demographic parity without levelling down. These demonstrations were prepared using the AutoGluon-Fair fairness toolkit.[152]

## 6.1   Example 1: Demographic parity

Figure 3 shows a Pareto Frontier for accuracy and demographic parity on the Adult Dataset. We show the results on the training set where demographic parity (Example 1) and true negative rate (Example 2) are enforced. Transferring them to the unseen test data introduces noise which would make the results less clear.

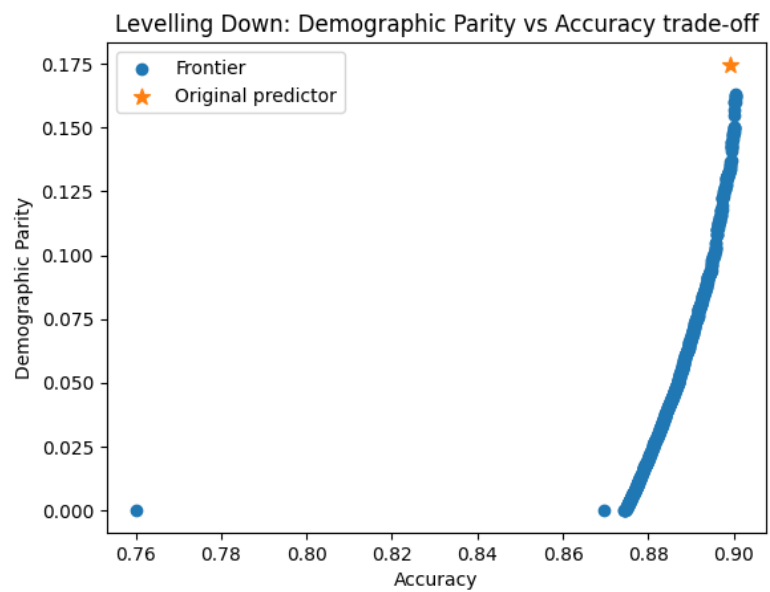

*Figure 3 - Tradeoff of demographic parity vs. accuracy when enforcing demographic parity in Example 1*

---

The dot on the far left represents a constant classifier that is perfectly fair. In Figure 4, we compute the selection rate per group for every classifier on the frontier. As expected, enforcing demographic parity exhibits levelling down with the selection rate for the advantaged group continually decreasing.

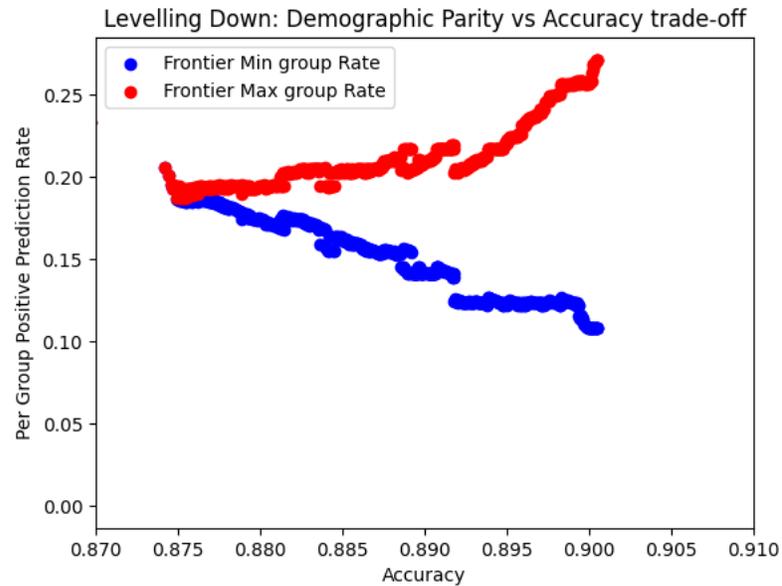

*Figure 4 - Tradeoff of positive prediction rate vs. accuracy when enforcing demographic parity in Example 1*

In contrast, if we instead enforce that the minimal selection rate for any group needs to be above a particular threshold, we observe very different behaviour as seen in Figure 5.

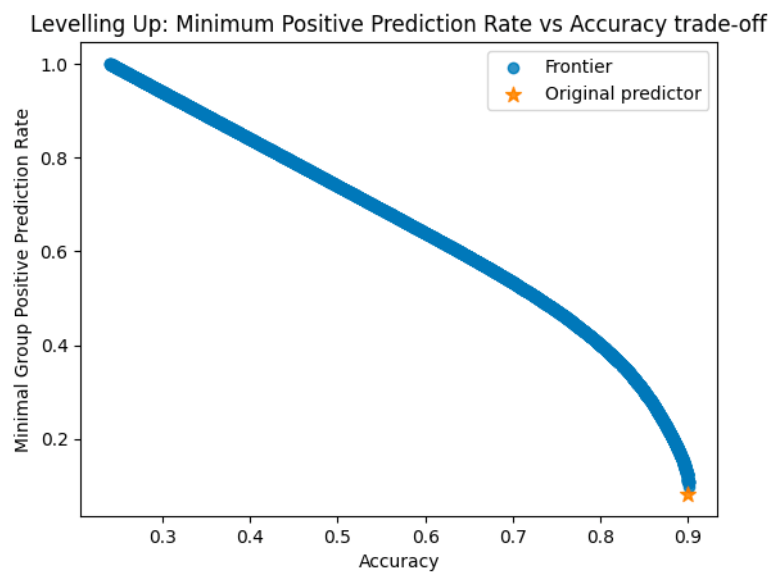

*Figure 5 - Tradeoff of minimum group positive prediction rate vs. accuracy when enforcing minimum positive prediction rate in Example 1*

As expected, with the dataset being more than 75% negatively labelled, large drops in accuracy were required for the positive decision rate to approach 1. Figure 6 shows the positive prediction rate for each group. Unlike enforcing egalitarian group fairness constraints, levelling down does not occur. Instead, the decision rate of the disadvantaged group steadily increases until it reaches parity with the advantaged group, followed by the decision rate for both groups increasing together.

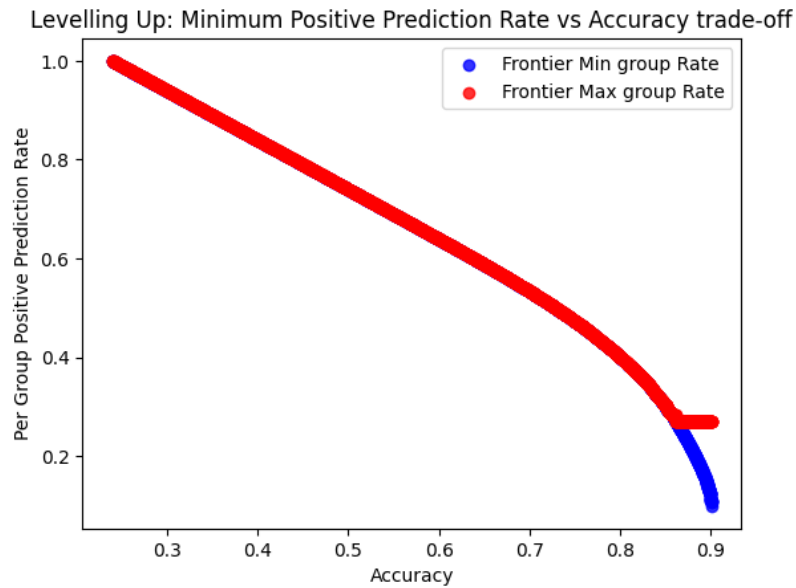

*Figure 6 - Tradeoff of per group positive prediction rate vs. accuracy when enforcing minimal positive prediction rate in Example 1*

As can be seen in the plots of the demographic parity for the frontier below (see: Figure 7), demographic parity is decreased, without levelling down, until parity is reached and then it is consistently near zero, as the selection rate increases for all groups.

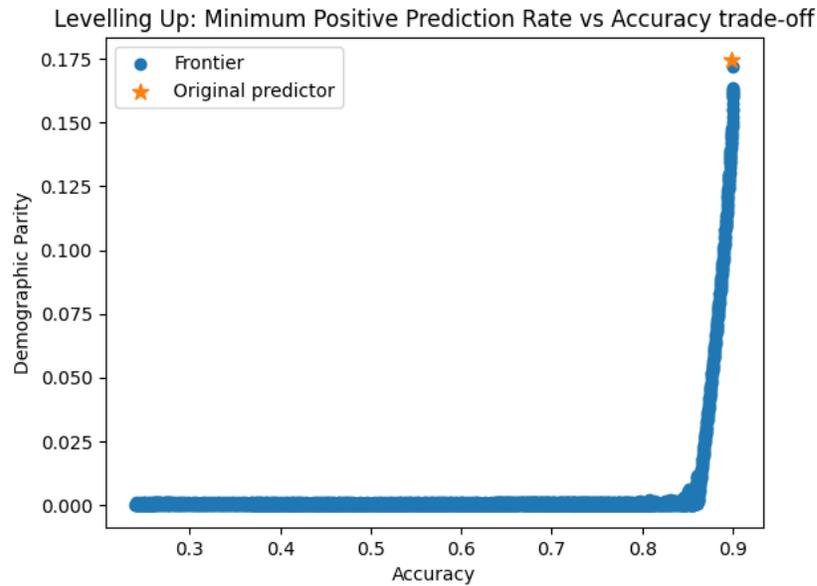

*Figure 7 - Tradeoff of demographic parity vs. accuracy when enforcing minimum positive prediction rate in Example 1*

## 6.2 Example 2: Difference in true negative rate

The same behaviour can be observed for other choices of equality metric. Figure 8 below shows the same behaviour for difference in true negative rate (or what is also called "false positive error rate balance" or "predictive equality" in [153]). Results for equal opportunity (i.e., difference in true positive rate) have similar behaviour, but owing to the small proportion of datapoints with a positive label, the frontier has less than 10 points, making the plot much less clear for our purposes.

---

[153] Verma and Rubin, *supra* note 14.

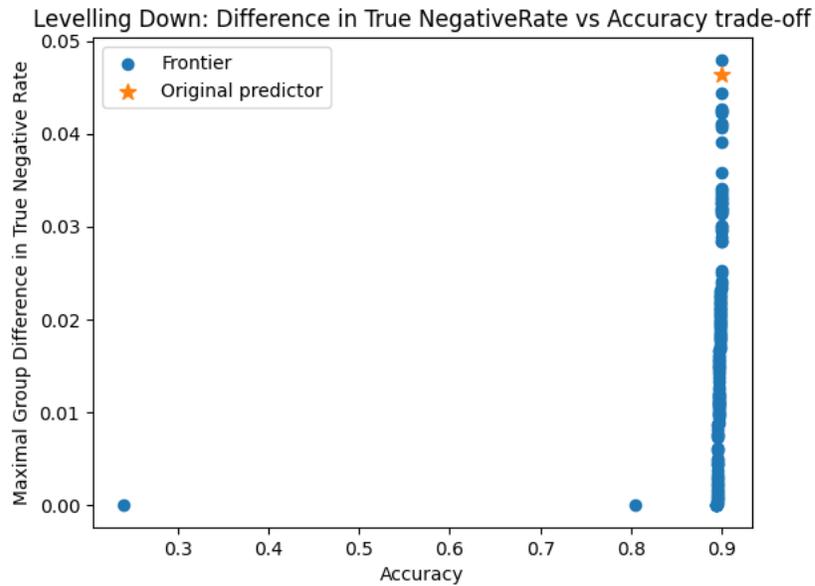

*Figure 8 - Tradeoff of true negative rate vs. accuracy when minimising difference in true negative rate in Example 2*

A close-up plot showing levelling down in the vertical component of the above frontier is shown below (see: Figure 9). While the curve is noisier than for demographic parity, a general trend of the true negative rate decreasing for the advantaged group is apparent.

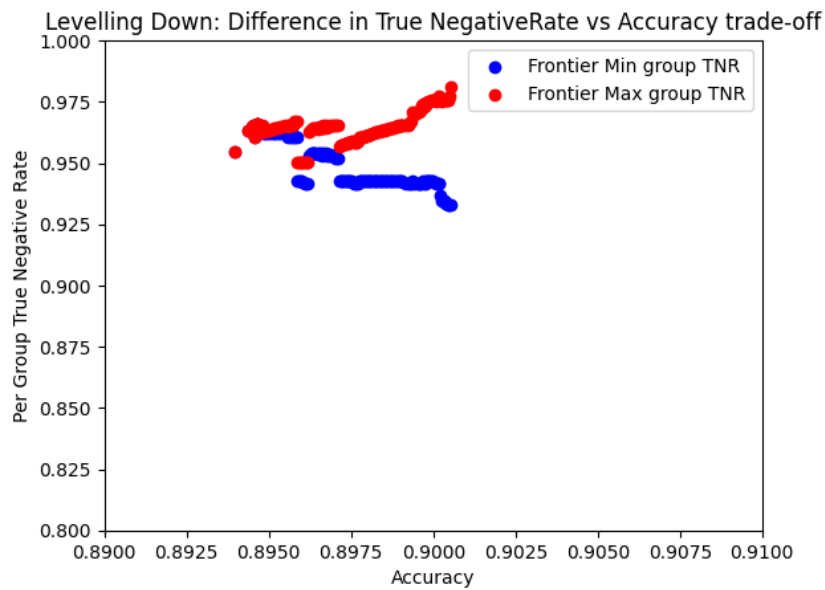

*Figure 9 - Tradeoff of per group true negative rate vs. accuracy when minimising difference in true negative rate in Example 2*

If instead we compute the frontier for minimum true negative rate vs accuracy, we see the following frontier in Figure 10.

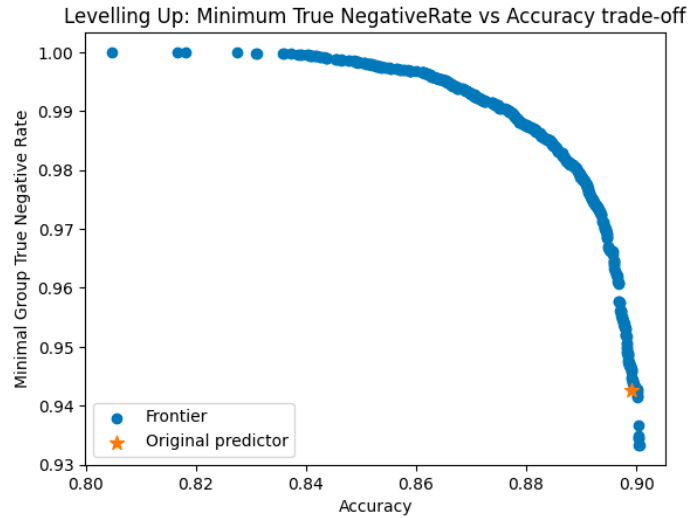

*Figure 10 - Tradeoff of minimum true negative rate vs. accuracy when maximising true negative rate in Example 2*

Plotting the per group response shows the following levelling up behaviour in Figure 11.

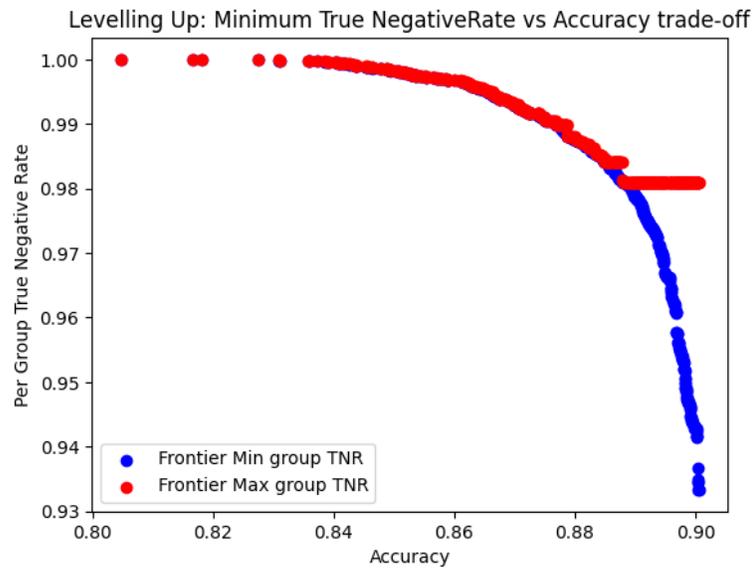

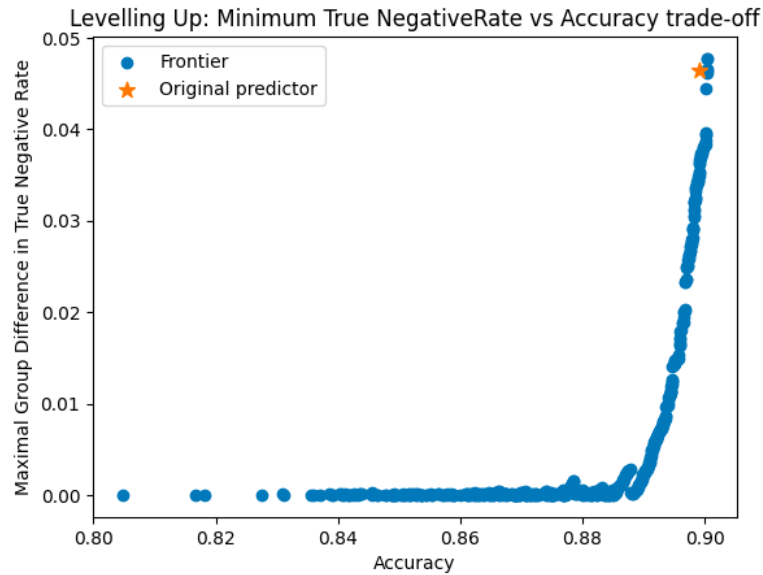

*Figure 11 - Tradeoff of per group true negative rate vs. accuracy when maximising true negative rate in Example 2*

This has the expected behaviour of decreasing the difference in true negative rate until the disadvantaged group has a near equivalent true negative rate to the advantaged group, after which the difference remains close to zero.

Directly comparing the behaviour of standard equality constraints with the minimum rate constraints (i.e., levelling up) plotted above reveals a few insights. First, for every possible inequality value, there is a corresponding choice of minimum rate threshold that will make the inequality smaller than, or of the same size, as this value. However, because these minimal value constraints cannot be satisfied by levelling down, a solution found by levelling up will typically have lower accuracy for a given equality level than solutions based on standard equality constraints. Nonetheless, MRCs can be informative with much smaller values than are needed to enforce standard equality constraints.

Taking this comparison further, in Figure 12 we compare levelling up with standard (or accuracy-maximizing) demographic parity on the Adult Dataset. The blue bars show the overall positive decision rate for an unconstrained classifier which makes positive decisions at a substantially lower rate for women than men. By enforcing demographic parity, standard fairness methods (see: Section 3.2) reduce harms to women at the group level but at the cost of increasing harms to men. In contrast, the green bars show an example of achieving demographic parity by levelling up the positive decision rate for women until parity is achieved without altering decision rates for men. Finally, the red bars show an example of partial levelling up (as discussed above) where the decision rate for women is improved to the same level enforced by standard demographic parity without also decreasing the decision rate or accuracy for men. It is interesting to note that only standard demographic parity results in a drop of accuracy for men, and that the results for women show that it is possible to substantially increase the positive decision rate for women while maintaining an accuracy above or comparable to men.

The green and red bars effectively represent a choice of different MRCs for positive decision rates. Maximising this rate (green bar) has a higher cost in terms of accuracy than partial levelling up (red bar), but achieves parity between men and women without levelling down. In contrast, partial levelling up retains more accuracy (even compared to demographic parity) but with a lower minimum positive decision rate threshold. Which of these MRCs is preferable or justifiable in a given use case or dataset is not immediately self-evident. But this is precisely the point—this type of normative decision cannot and should not be made solely from the perspective of what is easiest or most mathematically convenient.

Levelling up avoids the 'pathway of least resistance' by changing the goal of fairness from achieving strict parity between groups to ensuring a minimum 'fair' performance threshold is met. It helps guarantee that reductions in performance will only be used when they are causally necessary to improve performance for another group. It reframes fairness in machine learning as a question of harms rather than strict parity. By doing so, it necessitates local discussion to determine normatively acceptable levels of harm and to identify additional steps to be taken or resources to be allocated to meet them in practice. Bridging this gap between current practices and normative ideals is precisely where the real work of achieving substantive equality through fairML should begin.

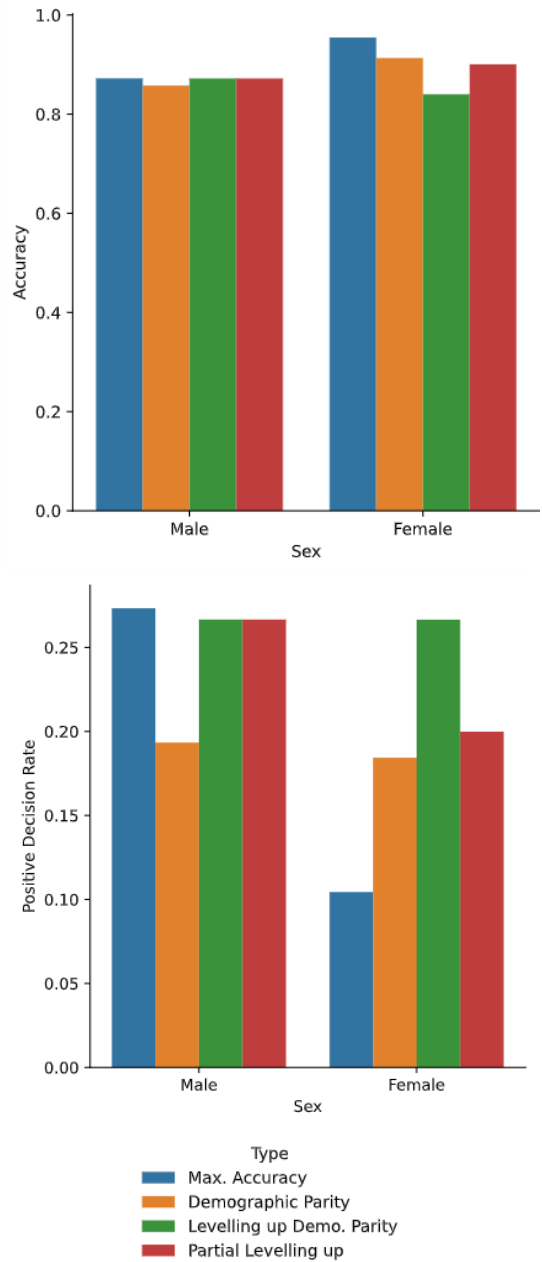

*Figure 12 - A comparison positive decision rates and accuracy when enforcing demographic parity either as an equality constraint or through levelling up*

## 7    CONCLUSION

Levelling down is a symptom of the choice to measure fairness solely in terms of disparity between groups and assume uniform value for benefits and harms, while ignoring welfare, priority, and other goods as well as stigmatisation, unequal concern, loss of solidarity, and other substantive harms which are central to questions of equality in the real world. Our examination of group fairness enforcement methods, philosophical theories, and equality law jurisprudence shows that levelling down is not a satisfactory solution to distributive justice problems in AI and ML. We call on researchers, developers, and deployers to engage seriously with the messy socioeconomic, legal, and philosophical details of the distributive justice problems to which fairML measures and methods are meant to be applied.

Substantively improving classifier performance, in comparison to levelling down, can be difficult, time and resource consuming, may require new data, and is not always possible for well-designed systems. Levelling down is nonetheless not the inevitable fate of enforcing fairness; rather, it is the result of taking the easier path out of mathematical convenience, and not any overarching societal, legal, or ethical reasons. Fairness cannot continue to be treated as a simple mathematical problem.

Moving forward, we see three possible pathways for fairML:

1. We can continue to deploy biased systems that ostensibly benefit only one privileged segment of the population while harming others.
2. We can continue to define fairness in formalistic mathematical terms and deploy AI and ML systems that perform worse for all groups and actively harmful for some groups.
3. We can take action and achieve fairness through "levelling up," meaning we design systems to purposefully generate more false positives for (historically) disadvantaged groups and dedicate the necessary additional resources to follow up with them more often (e.g., increased frequency of cancer screenings).[154]

Throughout this paper, we have outlined many reasons to reject levelling down. It shows unequal concern for disadvantaged groups, undermines social solidarity, and stigmatises. It causes unnecessary harm for advantaged groups in cases where a false negative can be incredibly costly in terms of health, welfare, and opportunities. But more than anything, it represents a missed opportunity to use AI and ML systems to live up to the substantive aims of equality and force reconsideration of deeply embedded inequality in the status quo.

In our view, only fairness achieved through levelling up is morally, ethically, and legally satisfactory. Levelling up is a more complex challenge: it needs to be paired with active steps to root-out the real life causes of biases in AI systems. Technical solutions are often only a plaster to deal with a broken system. But to fix the system, technical solutions need to be coupled with actions to achieve substantive equality. Improving access to healthcare, more diverse data sets, determining the true subjective value of benefits and harms to affected populations, and developing tools that are designed with disenfranchised communities in mind are some of the steps that have to be taken to do so.

---

[154] For example, modern definitions of algorithmic fairness, such as equal opportunity, can also be satisfied by "levelling up," or increasing the rate of cancer diagnosis until the recall is the same for every demographic group.

This is the challenge for the future of fairness in AI: to create systems that are substantively fair through levelling up, not only procedurally fair through levelling down. This is a much more complex challenge than simply tweaking a system to make two numbers equal between groups. It may require not only significant technological and methodological innovation, including re-designing AI systems from the ground up, but also substantial social changes in areas such as healthcare access and expenditures.

Difficult though it may be, this refocusing on fairness through levelling up is essential. AI systems make life changing decisions. Choices about how they should be fair, and to whom, are too important to reduce to a solvable mathematical problem. This is the status quo which has resulted in fairness methods that achieve equality through levelling down.

This is not enough. Existing tools are treated as a "solution" to algorithmic fairness, but thus far they do not deliver on their promise. Their morally murky effects make them less likely to be used and may be slowing down real solutions to these problems. We have created methods that are mathematically fair, but do not benefit the worst off. What we need are systems that are fair through levelling up, that help (historically) disadvantaged groups without arbitrarily harming others. This is the challenge fairML must now solve. We need AI that is substantively, not just mathematically, fair.

## APPENDIX 1: HARMS AND REMEDIES FOR GROUP FAIRNESS MEASURES

| Fairness measures | Justified use | Example | Direct harm to individuals | Direct remedy |
|---|---|---|---|---|
| *(Conditional) Demographic Parity (or statistical parity)* | Situations where historic data is expected to be prejudicial, and there is no agreed upon ground-truth. | Hiring, offering loans, access to education, representation in the media. | Lack of selection. | Increase or decrease selection rate. |
| *Equal Opportunity (or False negative error rate balance)* | Situations where there is agreed up on ground-truth and the overwhelming harm comes from false negatives. | Cancer or other serious illness screening. | Failure to identify positive cases. | Increase the recall. |
| *Predictive Parity** | Situations where there is agreed up on ground-truth and the overwhelming harm comes from false positives. | Misidentification as a known person of interest to the police. | Failure to identify negative cases. | Increases the precision. |
| *False positive error rate balance** | Situations where there is agreed up on ground-truth and the overwhelming harm comes from false positives. | Misidentification as a known person of interest to the police. | Failure to identify negative cases. | Increase the true negative rate. |
| *Equalized odds** | Combination of Equal Opportunity and False positive rate. | Treatment of illness by performing risky surgery. | Harms exist for failure to correctly identify positive and negative cases but they cannot be directly compared. | Increase recall and true positive rate simultaneously (may not be possible). |
| *Conditional use accuracy equality** | Combination of predictive parity and false positive error rate balance. | Treatment of illness by performing risky surgery. | Harms exist for failure to correctly identify positive and negative cases but they cannot be directly compared. | Increase precision and specificity simultaneously (may not be possible). |
| *Overall accuracy equality* | Situations where there is agreed up on ground-truth and the harm of misclassification is the same regardless of how people are situated. | Offering someone left- or right-handed scissors. | Harms exist for failure to correctly identify positive and negative cases and they are the same in both cases. | Increase overall accuracy simultaneously (may not be possible). |
| *Treatment Equality* | Unclear | - | -- | - |

* Predictive Parity and False Positive error rate balance both treat false negatives as a harm, but they normalize the harm differently. Predictive Parity is analogous to measuring the proportion of people in jail that are innocent, while False Positive Error Rate is analogous to measuring the proportion of innocent people that are in jail. While both measures relate to the number of innocent people in jail, whether this is recorded as a proportion of the people in jail, or of the total number of innocent people can drastically change what is seen as a significant harm. A similar relationship occurs between Equalized odds and Conditional Use Accuracy. Both are concerned with the same harms, but the way they are normalized varies. All fairness measures based on the classification scheme of Sahil Verma & Julia Rubin, *Fairness definitions explained*, in 2018 IEEE/ACM INTERNATIONAL WORKSHOP ON SOFTWARE FAIRNESS (FAIRWARE) 1 (2018).